\documentclass{article}

\usepackage[final,dandb]{neurips_2025}

\usepackage[utf8]{inputenc} 
\usepackage[T1]{fontenc}    
\usepackage{hyperref}       
\usepackage{url}            
\usepackage{booktabs}       
\usepackage{amsfonts}       
\usepackage{nicefrac}       
\usepackage{microtype}      
\usepackage{xcolor}         
\usepackage{wrapfig}

\usepackage{microtype}
\usepackage{graphicx}
\usepackage{subcaption}
\usepackage{booktabs} 

\usepackage{hyperref}

\usepackage{amsmath}
\usepackage{amssymb}
\usepackage{mathtools}
\usepackage{amsthm}
\usepackage{enumitem}
\usepackage{multirow}
\usepackage{booktabs}

\usepackage[capitalize,noabbrev]{cleveref}

\usepackage{algorithm}
\usepackage{algorithmicx}
\usepackage{algpseudocode}
\usepackage{xspace}
\usepackage{csquotes}
\usepackage{listings}
\usepackage{xcolor}

\usepackage{multirow}
\usepackage{multicol}
\usepackage{enumitem}
\usepackage[most,skins,theorems]{tcolorbox}
\tcbset{
  aibox/.style={
    width=\linewidth,
    top=8pt,
    bottom=4pt,
    colback=blue!6!white,
    colframe=black,
    colbacktitle=black,
    enhanced,
    center,
    attach boxed title to top left={yshift=-0.1in,xshift=0.15in},
    boxed title style={boxrule=0pt,colframe=white,},
  }
}

\newtcolorbox{AIbox}[2][]{aibox,title=#2,#1}

\definecolor{lightgray}{gray}{0.95} 
\definecolor{darkblue}{rgb}{0,0,0.6} 
\definecolor{nvgreen}{cmyk}{50, 0, 100, 0}

\newcommand{\yu}[1]{#1}

\theoremstyle{plain}

\theoremstyle{definition}

\theoremstyle{remark}

\DeclareMathOperator*{\argmin}{arg\,min}
\newcommand{\FullModelName}{Nemotron-CLIMB}
\newcommand{\ModelName}{CLIMB}
\newcommand{\climblab}{\textsc{Nemotron-ClimbLab}}
\newcommand{\finaldatamix}{\textsc{Nemotron-ClimbMix}}

\usepackage[textsize=tiny]{todonotes}

\title{
\makebox[\textwidth][c]{\FullModelName: \underline{CL}ustering-based \underline{I}terative Data}\\
\makebox[\textwidth][c]{\underline{M}ixture \underline{B}ootstrapping for Language Model Pre-training}
}

\author{Shizhe Diao$^{1}$, Yu Yang$^{1}$, Yonggan Fu$^{1}$, Xin Dong$^{1}$, Dan Su$^{1}$, Markus Kliegl$^{1}$, Zijia Chen$^{1}$,\\ \bf Peter Belcak$^{1}$,  \bf Yoshi Suhara$^{1}$, \bf Hongxu Yin$^{1}$, \bf Mostofa Patwary$^{1}$, \bf Yingyan (Celine) Lin$^{2}$, \\ \bf Jan Kautz$^{1}$,  \bf Pavlo Molchanov$^{1}$ \\
  $^{1}$NVIDIA
  $^{2}$Georgia Institute of Technology
  \\
}

\begin{document}

\maketitle

\begin{abstract}
Pre-training datasets are typically collected from web content and lack inherent domain divisions. 
For instance, widely used datasets like Common Crawl do not include explicit domain labels, while manually curating labeled datasets such as The Pile is labor-intensive. 
Consequently, identifying an optimal pre-training data mixture remains a challenging problem, despite its significant benefits for pre-training performance.
To address these challenges, we propose \textbf{CL}ustering-based \textbf{I}terative Data \textbf{M}ixture \textbf{B}ootstrapping (\textbf{{\FullModelName}}), an automated framework that discovers, evaluates, and refines data mixtures in a pre-training setting. Specifically, {\FullModelName} embeds and clusters large-scale datasets in a semantic space and then iteratively searches for optimal mixtures using a smaller proxy model and a predictor. 
This strategy enables effective domain adaptation without relying solely on curated data. 
When continuously trained on 400B tokens with this mixture, our 1B model exceeds the state-of-the-art Llama-3.2-1B by 2.0\%.
Moreover, we observe that optimizing for a specific domain (e.g., Social Sciences) yields a 5\% improvement over random sampling.
Finally, we introduce {\climblab}, a filtered 1.2T-token corpus with 20 clusters for research, and {\finaldatamix}, a 400B-token compact dataset designed for efficient pre-training that delivers superior performance under an equal token budget.
We analyze the final data mixture, elucidating the characteristics of an optimal data mixture.
Our data is available \href{https://huggingface.co/collections/nvidia/climb-datasets-67e428bdb9aaced2acda191f}{here}.
\end{abstract}

\section{Introduction}

\setlength\intextsep{0pt}
\begin{wrapfigure}{r}{0.39\textwidth}
\begin{center}
\includegraphics[width=\linewidth, trim=0 0 0 70, clip]{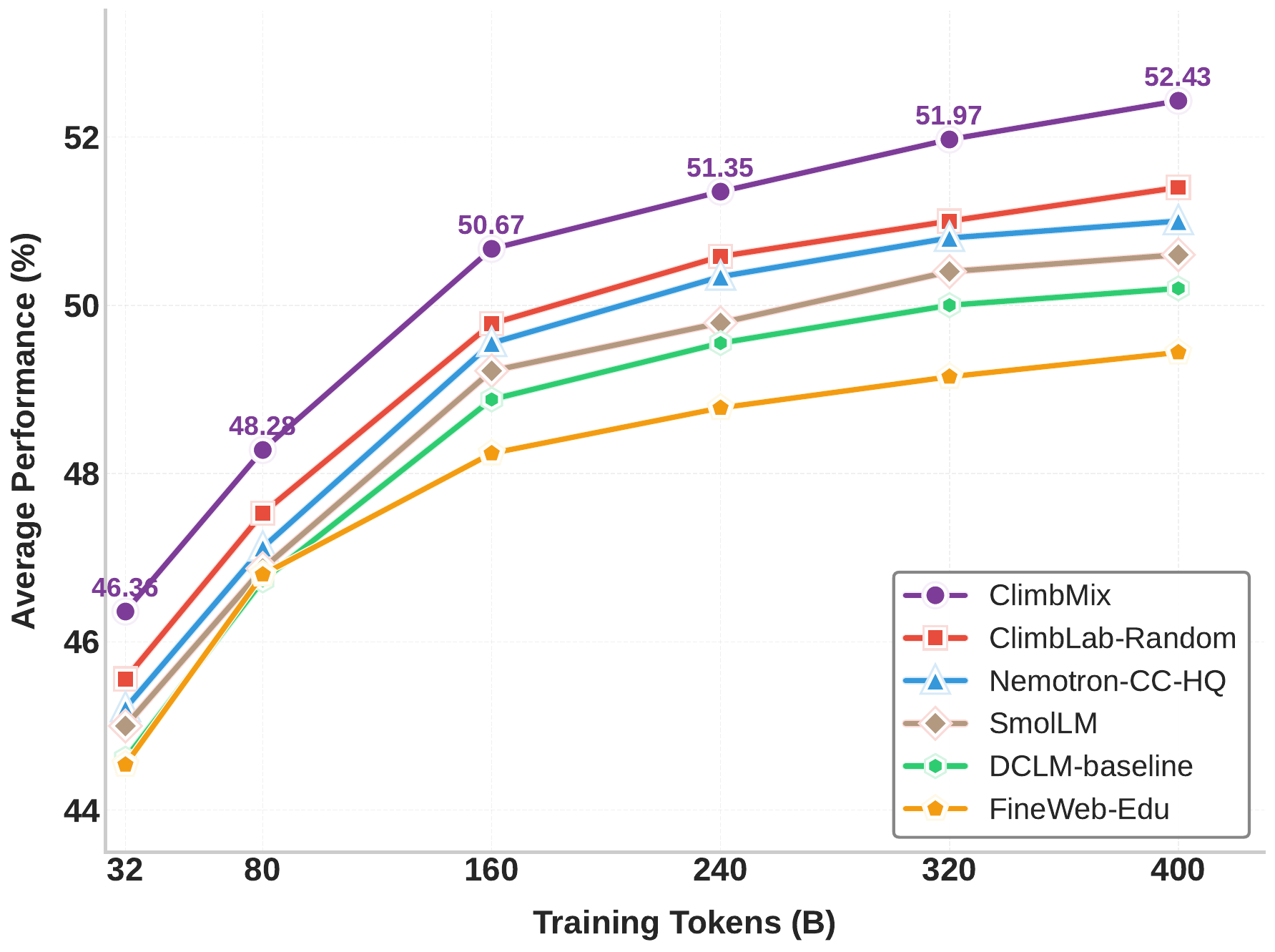}
\caption{    Pre-training a 1B model on {\finaldatamix} shows better scaling than training on other datasets. 
    We measure the average performance on 12 downstream benchmarks.}
\label{fig:final_data_comparison}
\end{center}
\end{wrapfigure}

Pre-training datasets for large language models (LLMs) have scaled to trillions of tokens, typically combining large-scale web crawls with smaller, high-quality domain-specific datasets. These corpora enable the development of generalist models capable of addressing diverse tasks. However, their vast scale and heterogeneity pose challenges in balancing general knowledge with domain expertise, often leading to inefficient utilization of high-value data for specialized capabilities.
Recent studies emphasize the importance of the final stage of pre-training, 
commonly referred to as mid-training, where models are refined on targeted, 
high-quality data to enhance specific capabilities.
For example, \cite{blakeney2024does} demonstrated that emphasizing domain-specific datasets during the final pre-training phase significantly improves performance on benchmarks such as GSM8K~\citep{cobbe2021training} (math), MMLU~\citep{hendryckstest2021} (reasoning), and HumanEval~\citep{chen2021evaluating} (coding). Similarly, OLMo 2~\citep{olmo20252olmo2furious} mixes high-quality web data with curated STEM references, synthetic math datasets, and encyclopedic content for mid-training, achieving notable gains in math reasoning tasks. These findings highlight the potential of carefully curated data mixtures in mid-training for improving domain performance.

\setlength\intextsep{0pt}
\begin{wrapfigure}{r}{0.39\textwidth}
\begin{center}
\includegraphics[width=0.9\linewidth, trim=0 0 0 70]{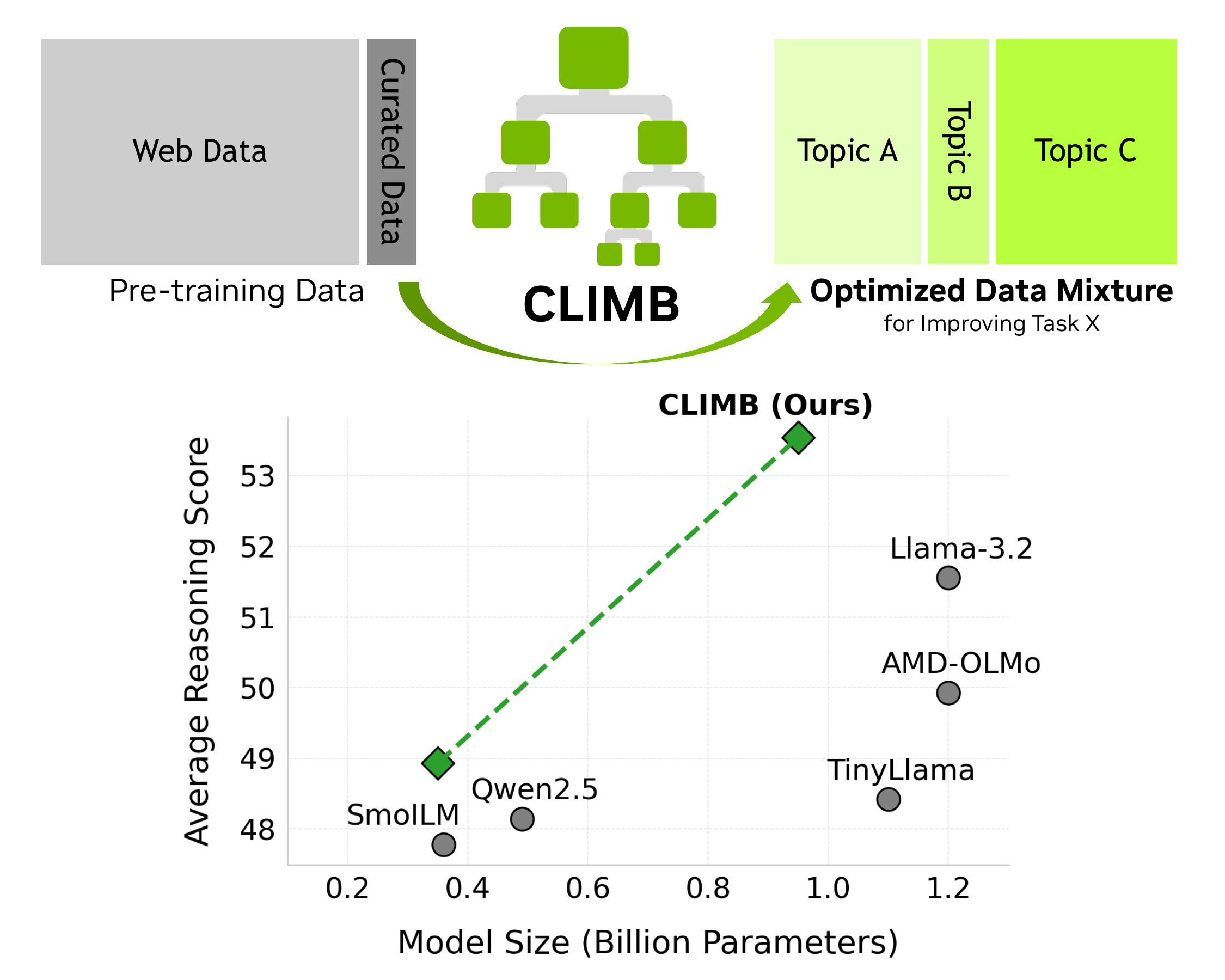}
\caption{Given large-scale pre-training data consisting of web-scale and curated sources, {\ModelName} identifies the optimal mixture of different topics (A, B, C) to improve performance in a target task (e.g., general reasoning). We compare the performance of state-of-the-art language models across different parameter scales on general reasoning benchmarks. {\ModelName} achieves a better tradeoff between model size and performance, demonstrating a more efficient scaling trend compared to prior models.}
\label{fig:concept}
\end{center}
\vspace{10pt}
\end{wrapfigure}

Despite the success of pre-training, optimizing data mixtures for both \textit{general} and \textit{domain-specific} tasks remains a challenge:
(1) Large-scale datasets such as Common Crawl~\footnote{\url{https://commoncrawl.org/}} offer unmatched diversity and scale but lack explicit domain labels, making it difficult to extract domain-relevant content. 
Filtering data often relies on general-purpose heuristics like perplexity or educational value~\citep{gunasekar2023textbooks}, which do not necessarily capture the most informative or high-quality content for specific domains.
(2) Even with curated datasets like The Pile~\citep{gao2020pile} with domain annotations, selecting an optimal data mixture is non-trivial due to the complex, nonlinear relationship between dataset composition and model performance.
For instance, optimizing a model for coding tasks requires not just programming-related content but also complementary knowledge from mathematics, reasoning, and security.

\setlength\intextsep{0pt}
\begin{wrapfigure}{r}{0.39\textwidth}
\begin{center}
\includegraphics[width=1.05\linewidth, trim=40 0 0 0, clip]{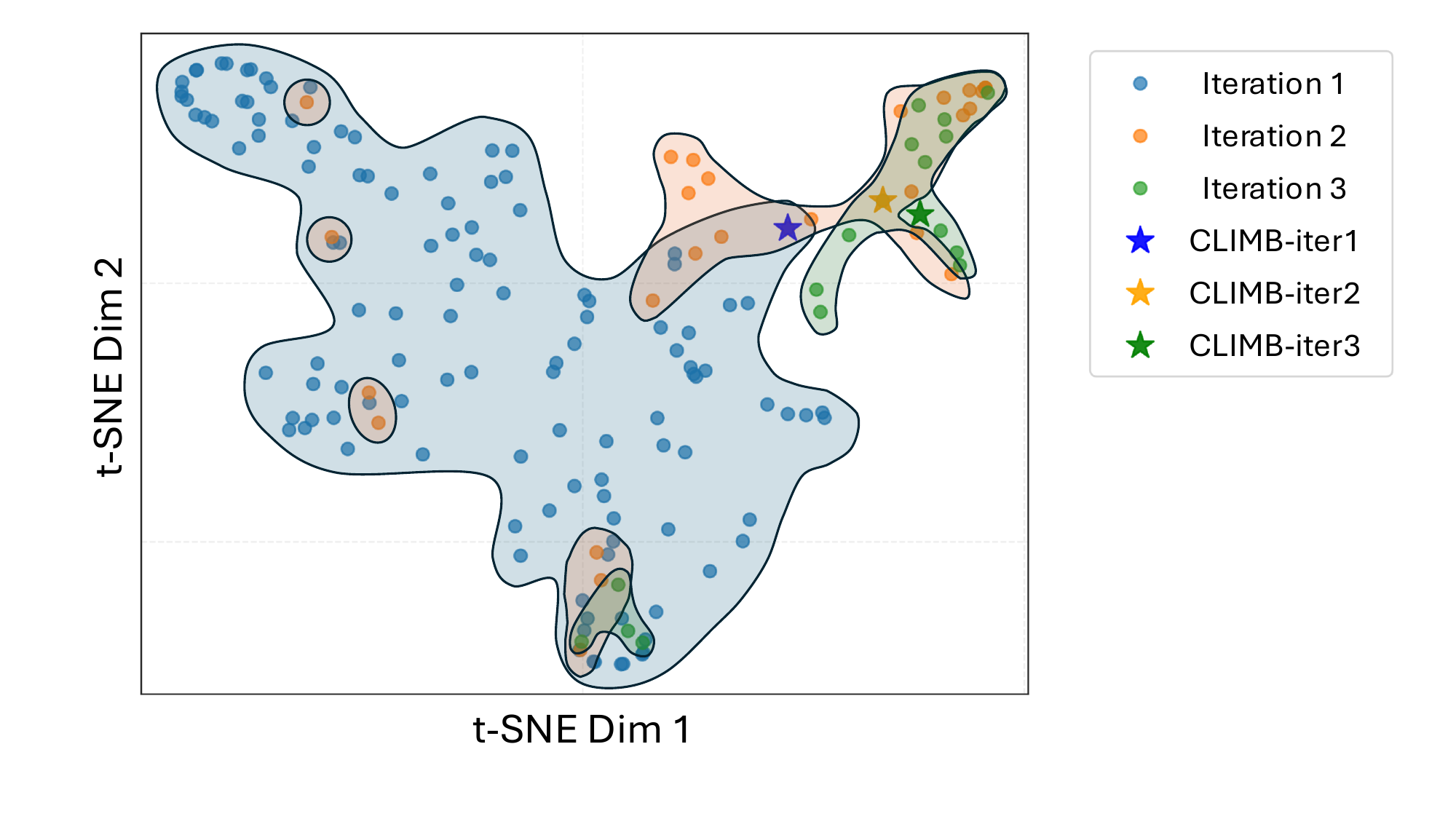}
\caption{Visualization of {\ModelName}'s iterative search process using t-SNE. 
Each point represents a data mixture config in the search space, with different iterations ({\ModelName}-Iter1, {\ModelName}-Iter2, {\ModelName}-Iter3) illustrating how the search space is refined over iterations. 
Initially, the search explores a broad set of configurations (Iter 1), progressively narrowing in subsequent iterations (Iter 2 and Iter 3) as {\ModelName} selects more optimal mixtures.}
\label{fig:tsne}
\end{center}
\end{wrapfigure}

To address these challenges, we propose \textbf{CL}ustering-based \textbf{I}terative Data \textbf{M}ixture \textbf{B}ootstrapping (\textbf{{\FullModelName}}; {\ModelName} for short)—a novel framework for automating the search for optimal pre-training data mixtures.
{\ModelName} consists of three key steps: (1) embedding and clustering large-scale datasets, (2) constructing mixture-performance pairs by sampling and pruning data mixtures and training proxy models, and (3) fitting a predictor.
By treating the data mixture as input features and performance metrics as target labels, we train a regression model as a predictor. 
This approach enables efficient, iterative refinement of data mixtures without relying on predefined domain labels.

We frame data mixture construction as a search problem and solve it using a bootstrapping strategy.
At each iteration, candidate mixtures are proposed, pruned, and refined to optimize diversity and domain relevance.
Unlike static mixing strategies, our method dynamically adjusts data mixtures throughout training using a weak predictor approach, integrating multiple predictors iteratively to discover effective configurations for domain adaptation.
{\ModelName} actively learns to refine and optimize data mixtures based on real-world feedback from environment verifications, rather than passively relying on predefined heuristics or human-annotated domain labels.
This ability to iteratively self-improve makes {\ModelName} more flexible and adaptive to new data distributions and domain-specific requirements.

Additionally, {\ModelName} prioritizes computational efficiency, demonstrating that iterative data mixture search achieves superior results within a fixed training budget.
For instance, rather than allocating all resources to a larger model searching in one iteration, our approach iteratively refines training data mixtures, balancing verification and generation tasks.
Importantly, to reduce the computational overhead, our method leverages lightweight proxy models to evaluate mixture quality and reduce the search space by pruning progressively, significantly reducing the cost of brute-force hyperparameter sweeps.

We demonstrate the effectiveness of {\ModelName} by searching the optimal data mixture in general reasoning tasks first and then extending it to specific domains (e.g., STEM, social sciences, and humanities).
Using the optimal data mixture discovered by {\ModelName}, we train 350M and 1B models on 40B tokens, both of which surpass the previously best data mixing (Doremi and RegMix) methods by a large margin. 
Furthermore, when trained on a larger number of tokens (400B) with this mixture, our 1B model exceeds the state-of-the-art Llama-3.2-1B by 2.0\%. 
We observe that optimizing for a specific domain (e.g., Social Sciences) yields a 5\% improvement over random sampling.
Finally, based on the insights obtained from our explorations, we further apply {\ModelName} to two existing datasets, Nemotron-CC~\citep{su2024nemotron} and smollm-corpus~\citep{benallal2024smollmcorpus}, and produce a new dataset with superior performance.

Our contributions are threefold:
\begin{itemize}[leftmargin=*,label=$\bullet$,noitemsep,partopsep=0pt,topsep=0pt,parsep=0pt]
    \item \textit{Automated Data Mixture Optimization.} We propose an embedding-driven data mixing approach to automatically identify, group, mix high-quality clusters, enabling domain-specific training while removing the reliance on manually predefined domain-specific data.
    \item \textit{Dynamic and Iterative Search Framework.} Our method introduces an iterative search process, dynamically refining data mixtures throughout training to optimize diversity and domain relevance, while addressing scaling challenges in clustering and data filtering.
    \item \textit{New High-quality Dataset.} We contribute a filtered 1.2-trillion-token corpus with 20 clusters as a new playground for data mixing research and a new high-quality 400-billion-token data for efficient pre-training.
\end{itemize}

\section{{\ModelName}: CLustering-based Iterative Data Mixture Bootstrapping}
\begin{figure}[t]
    \centering
    \includegraphics[width=1\textwidth]{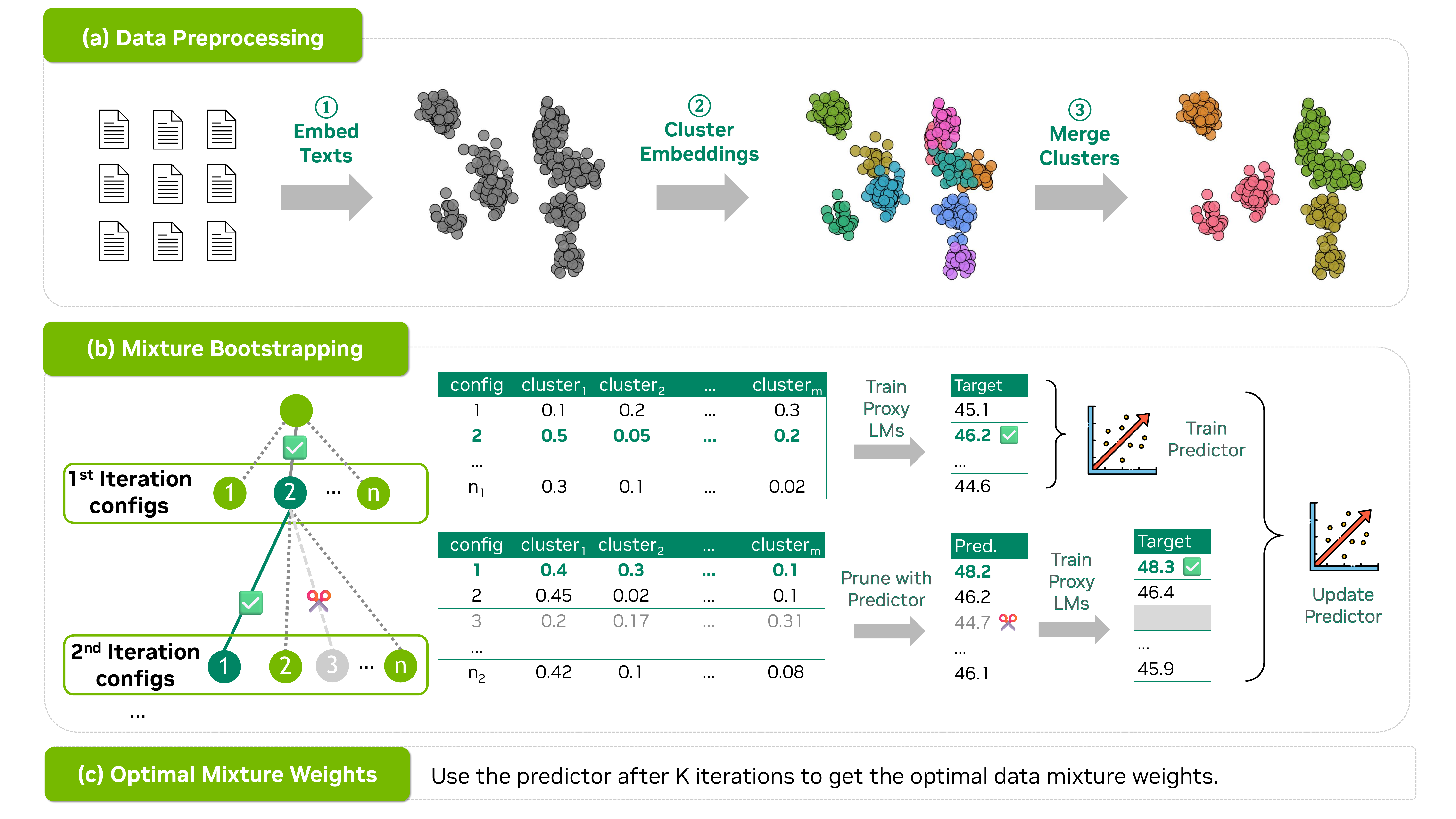}
    \caption{The {\ModelName} framework overview. \textbf{Upper section}: {\ModelName} first preprocesses raw data via embedding and clustering it into groups.
    These clusters serve as the basis for the search space, where a mixture is defined as a set of weights to combine different clusters. \textbf{Lower section}: 
    {\ModelName} samples $n_k$ mixtures in iteration $k$, trains proxy models on a subset of them, and updates a predictor to estimate performance. The predictor prunes mixtures that are likely to perform poorly, so only the most promising mixtures proceed to full proxy training in subsequent iterations. Through progressively refining the search space and eliminating suboptimal candidates, {\ModelName} converges toward an optimized data mixture and balances general and domain-specific performance without exhaustive manual curation.}
    \label{fig:model_arch}
    \vspace{-1 em}
\end{figure}

Our work focuses on curating training data from a massive data source in an automated fashion, specifically tailored to improve the desired tasks or domains. 
To ensure that the filtered dataset remains relevant to the target domain while maintaining general language modeling and reasoning capabilities, our framework simplifies the data curation process through a fully autonomous iterative bootstrapping approach, eliminating the need for manual curation and reducing labor costs.
As illustrated in Figure~\ref{fig:model_arch}, we first cluster documents from the data source in an embedding space to differentiate data across domains. Next, we iteratively optimize the mixture weights using a bootstrapping process to progressively enhance the dataset's domain relevance. Further details of the two phases are provided in Section~\ref{sec:clustering} and~\ref{sec:boostrapping}, respectively.

\subsection{Data Preprocessing}
\label{sec:clustering}

To effectively cluster documents belonging to the same domain, we propose clustering them in the embedding space rather than the word space, as this approach promotes a deeper semantic alignment among documents within the same cluster. To accomplish this, our framework follows three steps, as shown in Fig.~\ref{fig:model_arch} and elaborated below:

\textbf{Text embedding.}
Given a large raw dataset $\hat{D} = \{D_1, D_2, \dots, D_n\}$ containing $n$ documents, we map the documents into an embedding space using an embedding model $M_e$. The output of the embedding model is a set of embedding vectors, $E = \{E_1, E_2, \dots, E_n\}$.

\textbf{Embedding clustering.}
We then cluster the embeddings using a suitable clustering algorithm. 
For instance, k-means~\cite{hartigan1979algorithm} can be used to group them into $K_\text{init}$ clusters. 
To ensure the clusters are as fine-grained as possible for subsequent processing, we prefer to set $K_\text{init}$ to a relatively large value at this stage, such as $1000$. The specific settings are detailed in Section~\ref{sec:implementation}.

\textbf{Cluster merging.} 
To further improve clustering quality, we perform cluster pruning and merging. 
Specifically, given $K_\text{init}$ clusters, we conduct cluster-level pruning to remove low-quality clusters, retaining $K_\text{pruned}$ high-quality clusters based on model-based classifiers as the pruning metric. 
Then we merge the clusters into $K_\text{enhanced}$ clusters according to the distance between centroids, where $K_\text{enhanced} < K_\text{pruned} < K_\text{init}$.
The primary goal of merging is to merge similar fine-grained clusters and reduce the number of domains, facilitating the subsequent data mixture process. 
The entire dataset is reduced to $D$ from $\hat{D}$. 
The implementation details can be found in Section~\ref{sec:implementation}.
 
\subsection{Iterative Bootstrapping: Mixture Weight Search}
\label{sec:boostrapping}
Given a set of data clusters,
the next step is to optimize sampling mixture weights to maximize the desired downstream task performance. 
We formulate this as a bi-level optimization problem and solve it via iterative bootstrapping.

\textbf{Mixture weight search as a bi-level optimization problem.}
Given a set of data clusters \( D = \{D_1, D_2, \dots, D_k\} \) and the objective function \( \ell(\alpha, \omega) \) with model weights \( \omega \) trained with mixture weights \( \alpha \), which outputs the achievable performance \( P \) on a calibration set, the objective is to identify the optimal mixture weights \( \alpha^* \in A \) that maximize the task performance \( \ell(\alpha, \omega) \).
\begin{equation}
\min_{\alpha \in A} \,\, \ell_{val}(\alpha, \omega^*(\alpha)) 
\text{s.t.} \,\, \omega^*(\alpha) = \argmin_{\omega} \ell_{train}(\alpha, \omega) 
\text{s.t.} \sum_{i=1}^{k} \alpha_i = 1, \quad \alpha_i \geq 0
\end{equation}
\vspace{-1em}

\textbf{Approximate the objective with task performance predictors.}
A straightforward approach to estimate the objective function $\ell(\alpha, \omega)$ is to train a model for each $\alpha$ across the entire design space $A$. 
However, this is computationally prohibitive. 
To address this challenge, we propose using a predictor $f_{\theta}(\alpha)$ to approximate $\ell(\alpha, \omega)$ based on a subset of (\textit{mixture weights}, \textit{performance}) pairs, thereby significantly reducing the training cost.
In essence, our cluster mixture search can be reformulated as a bi-level optimization problem under the above approximation:
%

\begin{align}
\begin{split} \label{eq:approx}
\min_{\alpha \in A} f(\alpha | S)\,\, \textrm{s.t.} \,\, f &= \argmin_{S, f \in \tilde{\mathcal{F}}} \sum_{s \in S}\mathcal{L}(f(s), \ell(s, w^*))
\end{split}
\end{align}
where $\mathcal{L}$ is the loss function for the predictor $f_{\theta}$, $\tilde{\mathcal{F}}$ represents the set of all possible approximations to $\ell$, and $S := \{S \subseteq A \mid |S| \leq C\}$ denotes all configurations that satisfy the sampling budget $C$. The value of $C$ is directly tied to the total training cost of the proxy models.

\textbf{Iterative bootstrapping to solve Eq.~\ref{eq:approx}.} 
To solve Equation~\ref{eq:approx}, previous methods typically approach this optimization by first uniformly sampling mixture weights from the design space, training a model on the corresponding combined datasets, and then learning a predictor based on the performance of the trained models. 
However, we observe that, given a fixed training budget, this strategy is limited by the inefficiency of the initial uniform sampling. 
This inefficiency causes the model to focus excessively on low-quality mixture weights while failing to identify high-quality ones, ultimately leading to suboptimal mixture weights.

In light of this, rather than uniformly sampling across the entire space and then fitting the predictor, we propose an iterative approach to evolve both the sampling strategy $S$ and the predictor $f_{\theta}$. The rationale behind this method is to guide the predictor to focus more on subspaces with higher-quality weight mixtures, resulting in more accurate predictions under the same training budget. Specifically, this approach can be mathematically formulated as solving the bi-level optimization problem using a coordinate descent method that alternates between optimizing the configuration \emph{sampling} and predictor \emph{fitting} subroutines, where the iteration $k$ can be formulated as:
\begin{equation}
\begin{aligned}
&\textrm{(Sampling)} ~
&\tilde{P}^{k} = \{f_{k}(s) \vert s \in A\setminus S^{k} \}, ~
&S_M \subset \text{Top}_N(\tilde{P}^{k}),~S^{k+1} = S_{M} \cup S^{k},
\end{aligned}
\end{equation}

\begin{equation}
\begin{aligned}
&\textrm{(Predictor Fitting)} ~
&\alpha^*= \argmin_{\alpha \in A} f(\alpha | S^{k+1}),~
&\text{s.t.}\,\, f_{k+1} = \argmin_{f_{k} \in \tilde{\mathcal{F}}} \sum_{s \in S^{k+1}}\mathcal{L}(f(s), \ell(s, \omega^*))
\end{aligned}
\end{equation}
where $\text{Top}_N(\tilde{P}^{k})$ represents the set of the top $N$ configurations, ranked according to the task performance $\tilde{P}^{k}$.
In contrast, existing methods~\citep{liu2024regmix} can be seen as running the above coordinate descent process \textit{for only a single iteration}, which is a special case of our more general framework.

\textbf{Implementation.}
The above coordinate descent solution is intuitive and straightforward to implement.
Suppose that the iterative method consists of $K$ iterations. Initialize $S^1$ by randomly sampling a few configurations from $A$ and training proxy models to obtain their performance. Then, for iterations $k = 2, \dots, K$, jointly optimize the sampling set $S^k$ and the predictor $f_{\theta}^k$ in an alternating manner:

\textbf{\textit{Subroutine 1: Configuration sampling}.}
At iteration $k+1$, sort all configurations in the weight space $A$ (excluding those already in $S^k$) according to their predicted performance $\tilde{P}^k$. 
Next, randomly sample $M$ new configurations from the top $N$ ranked configurations based on $\tilde{P}^k$ in order to balance exploitation and exploration. 
These newly sampled configurations, combined with $S^k$, form $S^{k+1}$.

\textbf{\textit{Subroutine 2: (Weak) predictor fitting}.} 
Train a predictor $f_{\theta}^{k+1}$ by minimizing the loss $\mathcal{L}$ using the sampled configurations in $S^{k+1}$. The learned predictor $f_{\theta}^{k+1}$ is then used to evaluate the configurations and generate the predicted performance $\tilde{P}^{k+1}$.

By alternating between these two procedures for a predefined number of iterations, one progressively refines the predictors and guides the sampling process toward subspaces with higher-quality mixture weights, thereby increasing the average quality of the searched mixture weights. At the same time, the promising samples in $S^{k+1}$ improve the prediction accuracy of the updated predictor $f_{\theta}^{k+1}$ for high-performing configurations, allowing for more accurate assessment of the sampled configurations' quality. Finally, one selects the best configuration predicted by the final predictor as the final data mixture weight.
For implementation, the predictor can be any regression model, such as linear regression, ridge regression, decision tree regression, or a multilayer perceptron. In our experiments, we use LightGBM~\citep{ke2017lightgbm}, which predicts the target value by learning an ensemble of decision trees.
More implementation details could be found in Section~\ref{sec:implementation}.

\section{Experimental Settings}
\label{sec:exp_setting}

\textbf{Data.} For training, we use Nemotron-CC~\citep{su2024nemotron} and smollm-corpus~\citep{benallal2024smollmcorpus} as the source dataset. 
CLIMB-clustering yields 21 super-clusters containing 800B tokens.
For evaluation, we test on reasoning benchmarks: PIQA~\citep{bisk2020piqa}, ARC\_C, ARC\_E~\citep{allenai:arc}, HellaSwag~\citep{zellers2019hellaswag}, WinoGrande~\citep{sakaguchi2021winogrande}, and SIQA~\citep{sap2019socialiqa}. We optimize using PIQA, ARC\_E, and HellaSwag validation data, then evaluate on test sets. 
LM-Evaluation harness~\citep{eval-harness} is used, with all datasets in a 0-shot setting except MMLU (5-shot)~\citep{allal2024SmolLM,smollmevalsetup}.

\textbf{Model.} We first perform phase-1 pre-training to establish a solid foundation. Three Transformer decoder-only models (62M, 350M, 1B) are trained with next-token prediction on 10T tokens (a combination of DCLM~\citep{li2024datacomp} and TxT360~\citep{txt360data2024}), similar to~\citep{yang2024qwen2technicalreport} (12T tokens).
We use the warmup-stable-decay (WSD) learning rate schedule~\citep{hu2024minicpm}, allowing resumption in the stable stage and focusing on data mixing research in the decay stage. 
For proxy models, we use 62M and 350M for efficiency. 
For target models, we evaluate all three sizes to assess the approach across scales. 
For the rest of paper we use the 350M-proxy, ablations with 62M are in the Appendix~\ref{appendix:62m-proxy}.
Once the optimal data mixture is found, we train the target model on 40B tokens using this mixture and compare performance. 
Unless stated otherwise, all reported results come from this 40B continuous pre-training.

\textbf{Baselines.} We compare our method with (1) Random selection, and state-of-the-art data mixing methods, including (2) DoReMi~\citep{xie2024doremi}, and (3) RegMix~\citep{liu2024regmix}. 
The details about these baselines are in Appendix~\ref{appendix:baselines}.

\subsection{Implementation Details}

\label{sec:implementation}
\textbf{Text embedding.} We use \text{stella\_en\_400M\_v5}~\footnote{\url{https://huggingface.co/NovaSearch/stella_en_400M_v5}}, as it efficiently encodes large-scale text with excellent performance. 

\textbf{Embedding clustering.} We adopt the classic K-means clustering algorithm from the FAISS library~\citep{johnson2019billion, douze2024faiss}, setting the initial number of clusters $K_\text{init}$ to 1000.

\textbf{Cluster merging.} We train several fasttext models~\citep{joulin2016fasttext} to evaluate the data quality across four important dimensions - overall quality, educational value, informational value, and advertisement score (1-5) - by annotating 1 million texts with Nemotron-340B~\citep{adler2024nemotron} with a carefully designed prompt template (see Appendix~\ref{appendix:prompt}). 
Then we perform cluster-level pruning based on the fasttext scores, applying a relatively loose threshold of 3.0, which results in 240 (i.e., the value of $K_\text{pruned}$) clusters.
Finally, we group the clusters according to a Euclidean distance threshold of 1.5.

\textbf{Iterative bootstrapping.} The data mixture search runs for three iterations with 64, 32, and 16 searches in the first, second, and third iterations, respectively. 
We initialize a Dirichlet distribution based on each cluster’s token count and sample configurations. 
In each iteration, a predictor is trained using both current and past data.

For predictor training, we use a LightGBM~\citep{ke2017lightgbm} regression model, which fits mixture-performance pairs well with limited data~\citep{liu2024regmix}. 
To prevent overfitting, we set L1 and L2 regularization, early stopping, a maximum depth of four, and require at least five samples per leaf. 
The ablation study of the above design choices is in Section~\ref{sec:analysis}.
Additionally, we employed a separate validation set and an early stopping mechanism, halting training after 20 rounds of no improvement.

\begin{table}[t]
\centering
\scriptsize
\setlength{\tabcolsep}{4pt} 
\caption{Comparison with data mixture methods.
All models are continuously trained on the same number of tokens (40B).
The best results are highlighted in \textbf{bold}.
Base refers to the model before training and serves as the starting point for all other models. 
We report perplexity for wiki and lambda, accuracy for arc\_e, winogrande, siqa, accuracy\_norm for piqa, arc\_c, hellaswag.
}
\begin{tabular}{c|cc|cc|ccccccc}
\toprule
{\textbf{Size}} & {\textbf{Model}} & {\textbf{Proxy}} & \textbf{wiki} & \textbf{lambda} & \textbf{piqa} & \textbf{arc\_c} & \textbf{arc\_e} & \textbf{hellaswag} & \textbf{winogrande} & \textbf{siqa} & \textbf{avg.} \\ 
\midrule
\multirow{6}{*}{350M} 
 & Base & - & 22.70 & 8.87 & 70.03 & 28.11 & 56.12 & 51.16 & 54.48 & 40.75 & 50.11 \\ 
 & Random  & - & 20.92 & 9.85 & 71.16 & 30.54 & 62.50 & 52.14 & 55.40 & 41.29 & 52.17 \\ 
 & Doremi  & 350M  & 19.41 & 10.39 & 70.29 & 33.53 & 66.41 & 52.25 & 55.95 & 41.86 & 53.38 \\ 
 & RegMix & 350M  & 20.93 & 10.32 & 71.92 & 33.42 & 66.12 & 53.69 & 55.27 & 42.23 & 53.78 \\ 
 & {\ModelName}  & 350M & 19.67 & 9.29 & \textbf{72.21} & \textbf{34.87} & \textbf{67.25} & \textbf{55.32} & \textbf{56.79} & \textbf{42.54} & \textbf{54.83} \\ 
\midrule
\multirow{5}{*}{1B} 
 & Base & - & 17.79 & 6.65 & 73.89 & 34.92 & 66.77 & 62.12 & 59.82 & 41.26 & 56.46 \\ 
 & Random  & -  & 17.82 & 6.53 & 74.05 & 37.12 & 70.24 & 62.90 & 60.77 & 42.48 & 57.93 \\ 
 & Doremi  & 350M  & 15.78 & 6.33 & 74.91 & 40.01 & 72.34 & 63.53 & 61.08 & 43.09 & 59.16 \\ 
 & RegMix  & 350M  & 16.19 & 6.62 & 75.22 & 40.42 & 71.32 & 64.73 & 62.33 & 42.22 & 59.37 \\ 
 & {\ModelName} & 350M & 15.96 & 6.44 & \textbf{75.78} & \textbf{40.98} & \textbf{72.97} & \textbf{66.01} & \textbf{63.32} & \textbf{43.37} & \textbf{60.41} \\ 
\bottomrule
\end{tabular}%
\label{tab:350m-proxy-general-reasoning}
\end{table}

\section{Experimental Results}

In this section, we will demonstrate the effectiveness of {\ModelName}.
Firstly, we compare the performance of {\ModelName} with other data mixture methods (Table \ref{tab:350m-proxy-general-reasoning}).
Then with the optimal data mixture, we train longer and compare the model with stage-of-the-art baseline models.
We use general reasoning tasks as the benchmark and a 350M proxy model in the main experiment.

\subsection{Comparison with Data Mixture Baselines}
As shown in Table~\ref{tab:350m-proxy-general-reasoning}, {\ModelName} outperforms all baseline data mixture methods.
For example, with the 350M target model, {\ModelName} achieves an average accuracy of 54.83\%, outperforming Random (52.17\%) and the best-performing baseline, Regmix (53.78\%).
Similarly, for the 1B model, {\ModelName} achieves an average accuracy of 60.41\%, higher than all baselines.
Although the optimization objective is confined to the validation sets of PIQA, ARC\_E, and HellaSwag, we observe that the resulting performance gains carry over to all the benchmark tasks. 
This clearly demonstrates the robust generalization ability of our approach, indicating that optimizing on a limited set of core tasks can effectively capture and transfer essential reasoning capabilities across a broader range of problems.

\begin{table}[t]
\centering
\setlength{\tabcolsep}{3pt} 
\caption{Comparison with state-of-the-art language models on general reasoning benchmarks. 
{\ModelName} is continuously trained on 400B tokens with the optimal data mixture.
Best results in \textbf{bold}.
}
\resizebox{\textwidth}{!}{%
\begin{tabular}{cc|cccccccccccccc}
\toprule
{\textbf{Model}} & {\textbf{Size}} & \textbf{piqa} & \textbf{arc\_c} & \textbf{arc\_e} & \textbf{hellaswag} & \textbf{winogrande} & \textbf{siqa} & \textbf{mmlu}  & \textbf{obqa} & \textbf{boolq} & \textbf{race} & \textbf{lambda} & \textbf{truthfulqa} & \textbf{Avg.} \\ 
\midrule
Qwen2.5  & 490M  & 69.96 & 32.42 & 64.60 & 52.14 & 56.59 & 44.22 & 33.03 & 35.20 & 62.29 & 34.93 & 52.51 & 39.74 & 48.14 \\
SmolLM  & 360M  & 71.49 & 36.00 & 70.08 & 53.52 & 56.75 & 41.20 & 32.98 & 37.60 & 55.29 & 34.74 & 45.76 & 37.93 & 47.78 \\ 
\textbf{{\ModelName} (Ours)}  & 350M & 72.52 & 35.07 & 67.38 & 56.27 & 57.93 & 42.88 & 33.28 & 36.60 & 62.29 & 33.39 & 52.62 & 36.86 & 48.93 \\ 
\midrule
TinyLlama  & 1.1B  & 73.29 & 30.12 & 60.31 & 59.19 & 59.12 & 40.63 & 31.60 & 36.00 & 57.83 & 36.46 & 58.84 & 37.60 & 48.42 \\ 
AMD-OLMo  & 1.2B  & \textbf{75.63} & 33.70 & 65.95 & 63.61 & 61.64 & \textbf{44.17} & 31.92 & 35.80 & 60.58 & 34.64 & 59.31 & 32.22 & 49.93 \\ 
Llama-3.2  & 1.2B  & 74.59 & 36.26 & 65.49 & 63.67 & 60.69 & 42.99 & 35.40 & 37.20 & 63.98 & \textbf{37.80} & \textbf{62.99} & 37.67 & 51.56 \\ 
\textbf{{\ModelName} (Ours)} & 950M  & 75.46 & \textbf{40.96} & \textbf{73.57} & \textbf{66.90} & \textbf{63.54} & 43.55 & \textbf{36.47} & \textbf{41.20} & \textbf{66.02} & 36.65 & 59.05 & \textbf{39.06} & \textbf{53.54} \\ 
\bottomrule
\end{tabular}%
}
\label{tab:sota-llms}
\end{table}

\begin{figure}[t]
    \centering
    \includegraphics[width=0.95\textwidth]{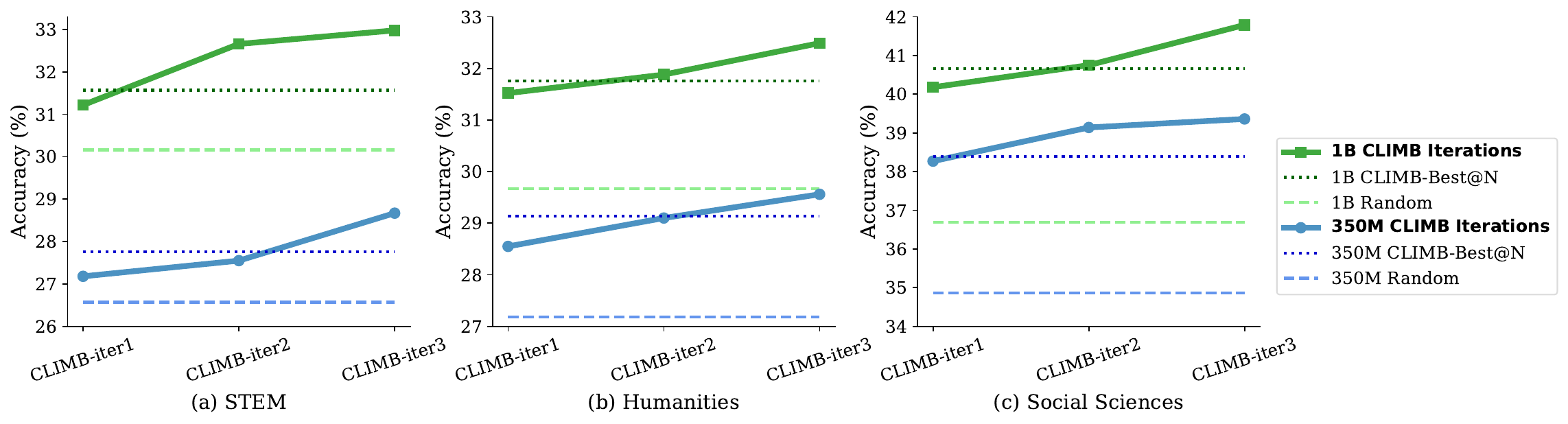}
    \caption{\yu{Performance of target models on MMLU benchmarks for different subject areas. 
    For both 350M and 1B target models, {\ModelName} used 350M proxy models, whereas {\ModelName}-Best@N used proxy models of the same size as the target models. {\ModelName} consistently improves performance across iterations, outperforming {\ModelName}-Best@N despite using smaller proxy models.}}
    \label{fig:mmlu}
\end{figure}

\subsection{Comparison with SOTA LMs}

Using the optimal data mixture identified by our method, we further investigate the effect of scaling up. 
Specifically, we used the same data mixture to train on 400B tokens and then compared the resulting model against state-of-the-art baselines.
As shown in Table~\ref{tab:sota-llms}, {\ModelName} achieves the best performance among all sub-500M and sub-1.2B models.
For example, when comparing models of similar scales (around 1B parameters), {\ModelName} consistently outperforms other baselines — including Llama-3.2 and AMD-OLMo — across the majority of the general reasoning benchmarks. 
In particular, it achieves the highest overall average score, surpassing the next-best model (i.e., Llama-3.2) by a noticeable margin (2.0\%). 
Moreover, we introduced additional benchmarks (e.g., mmlu, gpqa, obqa, boolq, and race), and our model is consistently better than baseline models, which demonstrates that our method exhibits excellent generalization performance.


\begin{AIbox}{Takeaway}
Iteratively refined data mixtures lead to better pre-training performance.
\end{AIbox}

\section{Analysis}

\label{sec:analysis}
In this section, we present the analysis and discussion about some important factors and designs behind {\ModelName} and demonstrate {\ModelName} is a robust data mixing method.

\textbf{Optimizating towards Specific Domains.}
In addition to optimizing towards general reasoning tasks, one important application of {\ModelName} is developing a domain-specialist model.
We explore searching the optimal data for specific domains.
Using the MMLU as an example, which has pre-defined three domains: STEM, humanities, and social-sciences and divided tasks into these domains, we conduct experiments on each domain separately.
We set the optimization objective as the performance on its validation set.
Here, we introduce a new baseline: {\ModelName}-Best@N, which directly searches for the best parameters on randomly sampled configs using the target model. 
Note that to ensure the same search compute in the table, the number of searches is reduced for the 1B model.
As shown in Figure~\ref{fig:mmlu}, the {\ModelName}-Best@N shows noticeably better accuracy than Random across all three domains, demonstrating the superiority of data searching.
This establishes a robust baseline for comparison. 
In contrast, our proposed {\ModelName} methods consistently improve performance across iterations. 
For instance, in the 350M model, {\ModelName}-iter3 achieves accuracies of 28.67\%, 29.56\%, and 39.36\% in STEM, Humanities, and Social Sciences, respectively, significantly outperforming both Random and {\ModelName}-Best@N. 
Similarly, in the 1B model, {\ModelName}-iter3 achieves a Social Sciences accuracy of 41.79\%, surpassing {\ModelName}-Best@N by 1.13\%. 
These results highlight the broad applicability of our approach to models of varying sizes.
In addition, we can see a clear improvement over each iteration. 
For example, from {\ModelName}-iter1 to {\ModelName}-iter3, the performance is improved from 40.18\% to 41.79\% on mmlu-social-sciences.

\begin{table}[t]
\centering
\footnotesize
\setlength{\tabcolsep}{4pt}
\caption{Ablation study with 1B target model trained on 40B tokens.
}
\resizebox{0.9\textwidth}{!}{%
\begin{tabular}{c|ccc|cccccccc}
\toprule
{\textbf{Setting}} & {\textbf{Model}} & {\textbf{Proxy}} & {\textbf{Comp.}} & \textbf{piqa} & \textbf{arc\_c} & \textbf{arc\_e} & \textbf{hellaswag} & \textbf{winogrande} & \textbf{siqa} & \textbf{Avg.} \\ 
\midrule
 \multirow{4}{*}{Abl.comp} 
 & {\ModelName}  & 350M & 100\% & 75.78 & 40.98 & 72.97 & 66.01 & 63.32 & 43.37 & 60.41 \\
 & {\ModelName}  & 350M & 150\% & 76.23 & 41.28 & 73.16 & 66.41 & 63.53 & 43.71 & 60.72 \\
 & {\ModelName} & 350M & 200\% & 76.51 & 42.31 & 73.41 & 66.81 & 63.70 & 43.99 & 61.12 \\
 \midrule
 \multirow{3}{*}{Abl.allo} 
 & {\ModelName}  & 350M & 6:1 & 75.32 & 40.80 & 72.91 & 65.51 & 62.84 & 42.93 & 60.05 \\
 & {\ModelName}  & 350M & 4:2:1 & 75.78 & 40.98 & 72.97 & 66.01 & 63.32 & 43.37 & 60.41 \\
 & {\ModelName}  & 350M & 2:2:1:1 & 75.36 & 40.77 & 72.88 & 65.86 & 62.97 & 43.02 & 60.14 \\
 \midrule
 \multirow{3}{*}{Abl.proxy} 
 & {\ModelName} & 62M & 100\% & 75.41 & 40.56 & 72.82 & 65.76 & 63.23 & 42.89 & 60.11 \\ 
 & {\ModelName} & 132M & 100\% & 75.56 & 40.93 & 72.94 & 65.57 & 63.09 & 43.07 & 60.19 \\
 & {\ModelName}  & 350M & 100\% & 75.78 & 40.98 & 72.97 & 66.01 & 63.32 & 43.37 & 60.41 \\
 \midrule
 \multirow{5}{*}{Abl.clus} 
 & 48-21cluster  & 350M & 100\% &  75.89 & 39.91 & 71.92 & 65.87 & 63.21 & 42.62 & 59.90 \\
 & 64-21cluster  & 350M & 100\% &  75.87 & 40.34 & 72.44 & 65.39 & 63.14 & 43.55 & 60.12 \\
 & 100-21cluster  & 350M & 100\% &  76.13 & 40.73 & 72.57 & 66.13 & 63.39 & 43.70 & 60.44 \\
 & 1000-21cluster  & 350M & 100\% & 75.78 & 40.98 & 72.97 & 66.01 & 63.32 & 43.37 & 60.41 \\
 & 2000-21cluster  & 350M & 100\% & 75.37 & 41.33 & 72.47 & 65.79 & 63.46 & 42.99 & 60.24 \\
 & 1000-15cluster  & 350M & 100\% & 75.94 & 41.33 & 73.34 & 66.28 & 63.62 & 43.05 & 60.59 \\
 & 1000-30cluster  & 350M & 100\% & 76.03 & 40.49 & 72.66 & 65.78 & 63.45 & 43.12 & 60.25 \\
 \midrule
 \multirow{2}{*}{Abl.init} 
 & Random  & 350M & 100\% & 75.42 & 40.12 & 72.47 & 65.73 & 64.27 & 43.22 & 60.21 \\ 
 & Dirichlet & 350M & 100\% & 75.78 & 40.98 & 72.97 & 66.01 & 63.32 & 43.37 & 60.41 \\
\bottomrule
\end{tabular}%
}
\label{tab:ablation-study}
\end{table}


\textbf{Effects of Search Compute Budget.} In the main experiments, we fix our total search budget (total compute) at 100\%.
Concretely, we perform three iterations of search with 64, 32, and 16 candidates evaluated in iterations 1, 2, and 3, respectively, giving a total of 112 searches.
To understand how scaling search computes helps, we compare runs with greater total numbers of searches (e.g., 168, 224).
Increasing the total number of searches allows the search procedure to more thoroughly explore possible data‐mixture candidates each iteration.
As shown in Table~\ref{tab:ablation-study} (rows under “Abl.comp”), we observe a trend that more extensive searches (e.g., 150\% or 200\%) continue to offer gains. 
This confirms our intuition that more exhaustive data‐mixture optimization can further boost downstream accuracy when sufficient compute is available.

\textbf{Effects of Compute Allocation.}
By default, we allocate our 100\% total compute across three iterations in a 4{:}2{:}1 ratio (64{:}32{:}16).
In principle, however, one could allocate compute to create either a “tall” search tree (more iterations but fewer searches per iteration) or a “fat” one (fewer iterations but more searches per iteration).
Table~\ref{tab:ablation-study} (rows under “Abl.allo”) compares several such allocations: 6{:}1, 4{:}2{:}1, and 2{:}2{:}1{:}1.
We find that 4{:}2{:}1 yields the best overall average performance (60.41\%).
Having too few iterations (e.g., 6{:}1) can lead to suboptimal exploration in earlier iterations, while splitting too many iterations (2{:}2{:}1{:}1) spreads compute too thin across each iteration.
Thus, balancing depth (number of iterations) and breadth (searches per iteration) proves key to robustly finding a good mixture.

\begin{AIbox}{Takeaway}
More search iterations improve performance, but compute should be balanced between depth and breadth. A 150\%-200\% compute increase yielded noticeable gains.
\end{AIbox}

\textbf{Analysis of Final Weights}
Furthermore, we analyzed the weights of the final data mixtures.
From Figure~\ref{fig:heatmap-combined} (a) , for the general reasoning task,
C8, C9, C18, and C19 account for the majority of the weight.
As shown in Appendix~\ref{appendix:relationship}, C8, C9, and C19 exhibit a high degree of correlation with general reasoning.
Moreover, when analyzing the topics of these four clusters (Table~\ref{tab:cluster_topics}), we find that they collectively form a diverse distribution.
More detailed analysis is shown in Appendix~\ref{appendix:analysis_of_final_weights}.

\begin{AIbox}{Takeaway}
Both the relevance of cluster content to downstream tasks and the diversity among different clusters are crucial for achieving effective data mixtures and robust model performance.
\end{AIbox}

In addition, we also discussed the topics of clusters, the relationship between clusters and downstream tasks, the effects of proxy models, the effects of the number of clusters, the effects of initialization, the effects of compute allocation, and the evolution of cluster weights in Appendix~\ref{appendix:analysis}.

\section{{\climblab} and {\finaldatamix}: New SOTA Pre-training Data}

Based on the insights obtained from our explorations above, we further apply {\ModelName} to two existing datasets: Nemotron-CC~\citep{su2024nemotron} and smollm-corpus~\citep{benallal2024smollmcorpus}, with the goal of constructing a powerful new pre-training dataset. 
Specifically, we first combine Nemotron-CC and smollm-corpus, and then employ our proposed \textbf{CLIMB-clustering} method to semantically reorganize and filter this combined dataset into 20 distinct clusters, leading to a 1.2-trillion-token high-quality corpus, named {\climblab}. 
Subsequently, we utilize \textbf{CLIMB-search} to identify an optimal data mixture from these clusters.
Using this optimal mixture, we further extract a 400-billion-token high-quality dataset named {\finaldatamix}.
We train a 1B model from scratch with {\finaldatamix} and evaluate its performance relative to models pretrained on other datasets under the same token budget. 
The results, illustrated in~Figure \ref{fig:final_data_comparison}, indicate that models trained on {\finaldatamix} significantly outperform those trained on existing datasets, including Nemotron-CC~\citep{su2024nemotron}, SmolLM~\citep{benallal2024smollmcorpus}, DCLM-baseline~\citep{li2024datacomp}, and FineWeb-Edu~\citep{penedo2024the}. 
The optimal data mixture weights identified by {\ModelName} is shown in Figure~\ref{fig:heatmap-clibmix}. 
We note that in the previous continuous pre-training setting, a few domains accounted for the majority of the weight. 
However, since the experiments here are conducted under a pre-training-from-scratch setting, a more balanced cluster distribution is required compared to continuous pre-training. 
This difference arises because continuous pre-training provides a strong foundation, allowing the model to focus primarily on learning a few important domains, whereas pre-training from scratch necessitates more diverse data coverage.
Finally, we publicly release these two datasets: the filtered 1.2-trillion-token dataset organized into 20 semantic clusters as a research playground for further data-mixture studies, and the optimized 400-billion-token {\finaldatamix} dataset for efficient pre-training.

\section{Related Work}

\textbf{Data Mixture for LLM Pre-training.} The composition of pre-training datasets are critical in determining the generalization abilities of language models ~\citep{radford2019language, brown2020language, touvron2023llama}. 
Typically, data mixtures like those in the Pile~\citep{gao2020pile}, GLaM~\citep{du2022glam}, and ROOTS~\citep{laurenccon2022bigscience} are crafted using manually defined rules, yet these heuristics lack standardization and transferability across different settings. 
SlimPajama-DC~\citep{shen2023slimpajama} systematically evaluated the influence of various predefined data configurations, yielding valuable insights. 
More recently, learning-based approaches such as DoReMi~\citep{xie2024doremi} and DoGE~\citep{fan2023doge} have introduced optimization techniques for domain proportions by iteratively refining training with reference and proxy models. 
While \citep{chen2024skill} investigated data sequencing strategies through the lens of curriculum learning, our work focuses on the simultaneous integration of diverse data domains, emphasizing a distinct aspect of pre-training.
The aforementioned methods show promise, but they require the dataset to already possess clear and natural domain distinctions. 
By contrast, we propose a novel approach that can automatically identify approximate domains from large amounts of web data and then find the optimal data mixture automatically.
In parallel work, WebOrganizer~\citep{wettig2025organize} proposes using classifiers to annotate web-scale data with topic and format labels. 
In contrast, our clustering-based approach is more straightforward, and readily scalable, and we introduce an iterative optimization method to refine the data mixture.

\textbf{Data Selection for Specific Domains.} 
Beyond optimizing the overall pre-training data mixture~\citep{zhang2024harnessing, yu2024mates, wettigqurating, brandfonbrener2024color}, selecting high-quality domain-specific data~\citep{ruder2017learning, xie2024efficient} is essential for improving model specialization during pre-training. 
Existing methods approach this challenge differently. 
DSIR~\citep{xie2023data} estimates relevance using hashed n-grams and resamples data to better match target domain distributions. 
CRISP~\citep{grangier2025taskadaptive} clusters the generalist dataset and samples these clusters according to their frequencies in the smaller specialist dataset.
\citep{kang2024get} propose to select data that nudges the pre-training distribution closer to the target distribution.
Training dynamics-based selection leverages model learning behavior to guide data filtering, including S2L~\citep{yang2024smalltolarge}, which clusters data based on loss trajectories to prioritize domain-relevant examples, and LESS~\citep{xia2024less}, which selects instruction tuning data with the highest gradient similarity to a target task. 
Embedding-based filtering removes redundant~\citep{abbas2023semdedup} or low-quality data, with SCIP~\citep{yang2023decoding} applying synthetic corruptions for filtering and heuristic pruning~\citep{singh2024brevity} reducing noise from overrepresented long-text clusters. 
\citep{shum2025predictive} proposes to select the data on which model losses are predictive of downstream abilities.
While these approaches improve data quality for specialized domains, they often rely on predefined domain labels or heuristics, limiting their flexibility for large-scale pre-training. 
In contrast, our proposed framework, {\ModelName}, iteratively refines domain-relevant data mixtures without requiring explicit domain labels, making it more applicable to real-world pre-training data and easier to scale. 

\section{Conclusion}

This work introduces {\ModelName}, a novel clustering-based iterative mixture bootstrapping framework for optimizing data mixture for pre-training LLMs. 
{\ModelName} automates the discovery, evaluation, and refinement of data mixtures, improving large-scale pre-training with explicit targets. 
By leveraging unsupervised clustering, proxy model training, and a predictor, {\ModelName} efficiently navigates the vast search space of data compositions, enabling the construction of optimal domain-aware mixtures without reliance on predefined domain labels or extensive manual curation.
By training 350M and 1B models with the optimal data mixture searched by {\ModelName}, we achieve state-of-the-art performance across 12 reasoning tasks.
Our experiments demonstrate that intelligently balancing unstructured corpora with targeted domain data leads to significant performance gains under fixed computational budgets. 
Compared to conventional static mixing strategies, our iterative approach allows for dynamic refinement, preserving general capabilities while excelling in specialized domains. 
Our findings underscore the potential of data-driven optimization techniques in enhancing LLM efficiency, advancing domain-specialized training.




\bibliography{neurips_2025}

\begin{thebibliography}{10}

\bibitem{blakeney2024does}
Cody Blakeney, Mansheej Paul, Brett~W Larsen, Sean Owen, and Jonathan Frankle.
\newblock Does your data spark joy? performance gains from domain upsampling at the end of training.
\newblock {\em arXiv preprint arXiv:2406.03476}, 2024.

\bibitem{cobbe2021training}
Karl Cobbe, Vineet Kosaraju, Mohammad Bavarian, Mark Chen, Heewoo Jun, Lukasz Kaiser, Matthias Plappert, Jerry Tworek, Jacob Hilton, Reiichiro Nakano, et~al.
\newblock Training verifiers to solve math word problems.
\newblock {\em arXiv preprint arXiv:2110.14168}, 2021.

\bibitem{hendryckstest2021}
Dan Hendrycks, Collin Burns, Steven Basart, Andy Zou, Mantas Mazeika, Dawn Song, and Jacob Steinhardt.
\newblock Measuring massive multitask language understanding.
\newblock {\em Proceedings of the International Conference on Learning Representations (ICLR)}, 2021.

\bibitem{chen2021evaluating}
Mark Chen, Jerry Tworek, Heewoo Jun, Qiming Yuan, Henrique Ponde De~Oliveira Pinto, Jared Kaplan, Harri Edwards, Yuri Burda, Nicholas Joseph, Greg Brockman, et~al.
\newblock Evaluating large language models trained on code.
\newblock {\em arXiv preprint arXiv:2107.03374}, 2021.

\bibitem{olmo20252olmo2furious}
Team OLMo, Pete Walsh, Luca Soldaini, Dirk Groeneveld, Kyle Lo, Shane Arora, Akshita Bhagia, Yuling Gu, Shengyi Huang, Matt Jordan, Nathan Lambert, Dustin Schwenk, Oyvind Tafjord, Taira Anderson, David Atkinson, Faeze Brahman, Christopher Clark, Pradeep Dasigi, Nouha Dziri, Michal Guerquin, Hamish Ivison, Pang~Wei Koh, Jiacheng Liu, Saumya Malik, William Merrill, Lester James~V. Miranda, Jacob Morrison, Tyler Murray, Crystal Nam, Valentina Pyatkin, Aman Rangapur, Michael Schmitz, Sam Skjonsberg, David Wadden, Christopher Wilhelm, Michael Wilson, Luke Zettlemoyer, Ali Farhadi, Noah~A. Smith, and Hannaneh Hajishirzi.
\newblock 2 olmo 2 furious, 2025.

\bibitem{gunasekar2023textbooks}
Suriya Gunasekar, Yi~Zhang, Jyoti Aneja, Caio C{\'e}sar~Teodoro Mendes, Allie Del~Giorno, Sivakanth Gopi, Mojan Javaheripi, Piero Kauffmann, Gustavo de~Rosa, Olli Saarikivi, et~al.
\newblock Textbooks are all you need.
\newblock {\em arXiv preprint arXiv:2306.11644}, 2023.

\bibitem{gao2020pile}
Leo Gao, Stella Biderman, Sid Black, Laurence Golding, Travis Hoppe, Charles Foster, Jason Phang, Horace He, Anish Thite, Noa Nabeshima, et~al.
\newblock The pile: An 800gb dataset of diverse text for language modeling.
\newblock {\em arXiv preprint arXiv:2101.00027}, 2020.

\bibitem{su2024nemotron}
Dan Su, Kezhi Kong, Ying Lin, Joseph Jennings, Brandon Norick, Markus Kliegl, Mostofa Patwary, Mohammad Shoeybi, and Bryan Catanzaro.
\newblock Nemotron-cc: Transforming common crawl into a refined long-horizon pretraining dataset.
\newblock {\em arXiv preprint arXiv:2412.02595}, 2024.

\bibitem{benallal2024smollmcorpus}
Loubna Ben~Allal, Anton Lozhkov, Guilherme Penedo, Thomas Wolf, and Leandro von Werra.
\newblock Smollm-corpus.
\newblock 2024.

\bibitem{hartigan1979algorithm}
John~A Hartigan and Manchek~A Wong.
\newblock Algorithm as 136: A k-means clustering algorithm.
\newblock {\em Journal of the royal statistical society. series c (applied statistics)}, 28(1):100--108, 1979.

\bibitem{liu2024regmix}
Qian Liu, Xiaosen Zheng, Niklas Muennighoff, Guangtao Zeng, Longxu Dou, Tianyu Pang, Jing Jiang, and Min Lin.
\newblock Regmix: Data mixture as regression for language model pre-training.
\newblock {\em arXiv preprint arXiv:2407.01492}, 2024.

\bibitem{ke2017lightgbm}
Guolin Ke, Qi~Meng, Thomas Finley, Taifeng Wang, Wei Chen, Weidong Ma, Qiwei Ye, and Tie-Yan Liu.
\newblock Lightgbm: A highly efficient gradient boosting decision tree.
\newblock {\em Advances in neural information processing systems}, 30, 2017.

\bibitem{bisk2020piqa}
Yonatan Bisk, Rowan Zellers, Jianfeng Gao, Yejin Choi, et~al.
\newblock Piqa: Reasoning about physical commonsense in natural language.
\newblock In {\em Proceedings of the AAAI conference on artificial intelligence}, volume~34, pages 7432--7439, 2020.

\bibitem{allenai:arc}
Peter Clark, Isaac Cowhey, Oren Etzioni, Tushar Khot, Ashish Sabharwal, Carissa Schoenick, and Oyvind Tafjord.
\newblock Think you have solved question answering? try arc, the ai2 reasoning challenge.
\newblock {\em arXiv:1803.05457v1}, 2018.

\bibitem{zellers2019hellaswag}
Rowan Zellers, Ari Holtzman, Yonatan Bisk, Ali Farhadi, and Yejin Choi.
\newblock Hellaswag: Can a machine really finish your sentence?
\newblock {\em arXiv preprint arXiv:1905.07830}, 2019.

\bibitem{sakaguchi2021winogrande}
Keisuke Sakaguchi, Ronan~Le Bras, Chandra Bhagavatula, and Yejin Choi.
\newblock Winogrande: An adversarial winograd schema challenge at scale.
\newblock {\em Communications of the ACM}, 64(9):99--106, 2021.

\bibitem{sap2019socialiqa}
Maarten Sap, Hannah Rashkin, Derek Chen, Ronan LeBras, and Yejin Choi.
\newblock Socialiqa: Commonsense reasoning about social interactions.
\newblock {\em arXiv preprint arXiv:1904.09728}, 2019.

\bibitem{eval-harness}
Leo Gao, Jonathan Tow, Baber Abbasi, Stella Biderman, Sid Black, Anthony DiPofi, Charles Foster, Laurence Golding, Jeffrey Hsu, Alain Le~Noac'h, Haonan Li, Kyle McDonell, Niklas Muennighoff, Chris Ociepa, Jason Phang, Laria Reynolds, Hailey Schoelkopf, Aviya Skowron, Lintang Sutawika, Eric Tang, Anish Thite, Ben Wang, Kevin Wang, and Andy Zou.
\newblock A framework for few-shot language model evaluation, 07 2024.

\bibitem{allal2024SmolLM}
Loubna~Ben Allal, Anton Lozhkov, Elie Bakouch, Leandro von Werra, and Thomas Wolf.
\newblock Smollm - blazingly fast and remarkably powerful, 2024.

\bibitem{smollmevalsetup}
Cosmopedia/evaluation.
\newblock Cosmopedia/evaluation at main · huggingface/cosmopedia.

\bibitem{li2024datacomp}
Jeffrey Li, Alex Fang, Georgios Smyrnis, Maor Ivgi, Matt Jordan, Samir~Yitzhak Gadre, Hritik Bansal, Etash Guha, Sedrick~Scott Keh, Kushal Arora, et~al.
\newblock Datacomp-lm: In search of the next generation of training sets for language models.
\newblock {\em Advances in Neural Information Processing Systems}, 37:14200--14282, 2024.

\bibitem{txt360data2024}
Liping Tang, Nikhil Ranjan, Omkar Pangarkar, Xuezhi Liang, Zhen Wang, Li~An, Bhaskar Rao, Linghao Jin, Huijuan Wang, Zhoujun Cheng, et~al.
\newblock Txt360: A top-quality llm pre-training dataset requires the perfect blend, 2024.

\bibitem{yang2024qwen2technicalreport}
An~Yang, Baosong Yang, Binyuan Hui, Bo~Zheng, Bowen Yu, Chang Zhou, Chengpeng Li, Chengyuan Li, Dayiheng Liu, Fei Huang, Guanting Dong, Haoran Wei, Huan Lin, Jialong Tang, Jialin Wang, Jian Yang, Jianhong Tu, Jianwei Zhang, Jianxin Ma, Jianxin Yang, Jin Xu, Jingren Zhou, Jinze Bai, Jinzheng He, Junyang Lin, Kai Dang, Keming Lu, Keqin Chen, Kexin Yang, Mei Li, Mingfeng Xue, Na~Ni, Pei Zhang, Peng Wang, Ru~Peng, Rui Men, Ruize Gao, Runji Lin, Shijie Wang, Shuai Bai, Sinan Tan, Tianhang Zhu, Tianhao Li, Tianyu Liu, Wenbin Ge, Xiaodong Deng, Xiaohuan Zhou, Xingzhang Ren, Xinyu Zhang, Xipin Wei, Xuancheng Ren, Xuejing Liu, Yang Fan, Yang Yao, Yichang Zhang, Yu~Wan, Yunfei Chu, Yuqiong Liu, Zeyu Cui, Zhenru Zhang, Zhifang Guo, and Zhihao Fan.
\newblock Qwen2 technical report, 2024.

\bibitem{hu2024minicpm}
Shengding Hu, Yuge Tu, Xu~Han, Chaoqun He, Ganqu Cui, Xiang Long, Zhi Zheng, Yewei Fang, Yuxiang Huang, Weilin Zhao, et~al.
\newblock Minicpm: Unveiling the potential of small language models with scalable training strategies.
\newblock {\em arXiv preprint arXiv:2404.06395}, 2024.

\bibitem{xie2024doremi}
Sang~Michael Xie, Hieu Pham, Xuanyi Dong, Nan Du, Hanxiao Liu, Yifeng Lu, Percy~S Liang, Quoc~V Le, Tengyu Ma, and Adams~Wei Yu.
\newblock Doremi: Optimizing data mixtures speeds up language model pretraining.
\newblock {\em Advances in Neural Information Processing Systems}, 36, 2024.

\bibitem{johnson2019billion}
Jeff Johnson, Matthijs Douze, and Herv{\'e} J{\'e}gou.
\newblock Billion-scale similarity search with {GPUs}.
\newblock {\em IEEE Transactions on Big Data}, 7(3):535--547, 2019.

\bibitem{douze2024faiss}
Matthijs Douze, Alexandr Guzhva, Chengqi Deng, Jeff Johnson, Gergely Szilvasy, Pierre-Emmanuel Mazaré, Maria Lomeli, Lucas Hosseini, and Hervé Jégou.
\newblock The faiss library.
\newblock 2024.

\bibitem{joulin2016fasttext}
Armand Joulin, Edouard Grave, Piotr Bojanowski, Matthijs Douze, H{\'e}rve J{\'e}gou, and Tomas Mikolov.
\newblock Fasttext.zip: Compressing text classification models.
\newblock {\em arXiv preprint arXiv:1612.03651}, 2016.

\bibitem{adler2024nemotron}
Bo~Adler, Niket Agarwal, Ashwath Aithal, Dong~H Anh, Pallab Bhattacharya, Annika Brundyn, Jared Casper, Bryan Catanzaro, Sharon Clay, Jonathan Cohen, et~al.
\newblock Nemotron-4 340b technical report.
\newblock {\em arXiv preprint arXiv:2406.11704}, 2024.

\bibitem{penedo2024the}
Guilherme Penedo, Hynek Kydl{\'\i}{\v{c}}ek, Loubna~Ben allal, Anton Lozhkov, Margaret Mitchell, Colin Raffel, Leandro~Von Werra, and Thomas Wolf.
\newblock The fineweb datasets: Decanting the web for the finest text data at scale.
\newblock In {\em The Thirty-eight Conference on Neural Information Processing Systems Datasets and Benchmarks Track}, 2024.

\bibitem{radford2019language}
Alec Radford, Jeffrey Wu, Rewon Child, David Luan, Dario Amodei, Ilya Sutskever, et~al.
\newblock Language models are unsupervised multitask learners.
\newblock {\em OpenAI blog}, 1(8):9, 2019.

\bibitem{brown2020language}
Tom Brown, Benjamin Mann, Nick Ryder, Melanie Subbiah, Jared~D Kaplan, Prafulla Dhariwal, Arvind Neelakantan, Pranav Shyam, Girish Sastry, Amanda Askell, et~al.
\newblock Language models are few-shot learners.
\newblock {\em Advances in neural information processing systems}, 33:1877--1901, 2020.

\bibitem{touvron2023llama}
Hugo Touvron, Thibaut Lavril, Gautier Izacard, Xavier Martinet, Marie-Anne Lachaux, Timoth{\'e}e Lacroix, Baptiste Rozi{\`e}re, Naman Goyal, Eric Hambro, Faisal Azhar, et~al.
\newblock Llama: Open and efficient foundation language models.
\newblock {\em arXiv preprint arXiv:2302.13971}, 2023.

\bibitem{du2022glam}
Nan Du, Yanping Huang, Andrew~M Dai, Simon Tong, Dmitry Lepikhin, Yuanzhong Xu, Maxim Krikun, Yanqi Zhou, Adams~Wei Yu, Orhan Firat, et~al.
\newblock Glam: Efficient scaling of language models with mixture-of-experts.
\newblock In {\em International Conference on Machine Learning}, pages 5547--5569. PMLR, 2022.

\bibitem{laurenccon2022bigscience}
Hugo Lauren{\c{c}}on, Lucile Saulnier, Thomas Wang, Christopher Akiki, Albert Villanova~del Moral, Teven Le~Scao, Leandro Von~Werra, Chenghao Mou, Eduardo Gonz{\'a}lez~Ponferrada, Huu Nguyen, et~al.
\newblock The bigscience roots corpus: A 1.6 tb composite multilingual dataset.
\newblock {\em Advances in Neural Information Processing Systems}, 35:31809--31826, 2022.

\bibitem{shen2023slimpajama}
Zhiqiang Shen, Tianhua Tao, Liqun Ma, Willie Neiswanger, Zhengzhong Liu, Hongyi Wang, Bowen Tan, Joel Hestness, Natalia Vassilieva, Daria Soboleva, et~al.
\newblock Slimpajama-dc: Understanding data combinations for llm training.
\newblock {\em arXiv preprint arXiv:2309.10818}, 2023.

\bibitem{fan2023doge}
Simin Fan, Matteo Pagliardini, and Martin Jaggi.
\newblock Doge: Domain reweighting with generalization estimation.
\newblock {\em arXiv preprint arXiv:2310.15393}, 2023.

\bibitem{chen2024skill}
Mayee Chen, Nicholas Roberts, Kush Bhatia, Jue Wang, Ce~Zhang, Frederic Sala, and Christopher R{\'e}.
\newblock Skill-it! a data-driven skills framework for understanding and training language models.
\newblock {\em Advances in Neural Information Processing Systems}, 36, 2024.

\bibitem{wettig2025organize}
Alexander Wettig, Kyle Lo, Sewon Min, Hannaneh Hajishirzi, Danqi Chen, and Luca Soldaini.
\newblock Organize the web: Constructing domains enhances pre-training data curation.
\newblock {\em arXiv preprint arXiv:2502.10341}, 2025.

\bibitem{zhang2024harnessing}
Chi Zhang, Huaping Zhong, Kuan Zhang, Chengliang Chai, Rui Wang, Xinlin Zhuang, Tianyi Bai, Jiantao Qiu, Lei Cao, Ju~Fan, et~al.
\newblock Harnessing diversity for important data selection in pretraining large language models.
\newblock {\em arXiv preprint arXiv:2409.16986}, 2024.

\bibitem{yu2024mates}
Zichun Yu, Spandan Das, and Chenyan Xiong.
\newblock Mates: Model-aware data selection for efficient pretraining with data influence models.
\newblock {\em Advances in Neural Information Processing Systems}, 37:108735--108759, 2024.

\bibitem{wettigqurating}
Alexander Wettig, Aatmik Gupta, Saumya Malik, and Danqi Chen.
\newblock Qurating: Selecting high-quality data for training language models.
\newblock In {\em Forty-first International Conference on Machine Learning}.

\bibitem{brandfonbrener2024color}
David Brandfonbrener, Hanlin Zhang, Andreas Kirsch, Jonathan~Richard Schwarz, and Sham Kakade.
\newblock Color-filter: Conditional loss reduction filtering for targeted language model pre-training.
\newblock {\em Advances in Neural Information Processing Systems}, 37:97618--97649, 2024.

\bibitem{ruder2017learning}
Sebastian Ruder and Barbara Plank.
\newblock Learning to select data for transfer learning with bayesian optimization.
\newblock In {\em Proceedings of the 2017 Conference on Empirical Methods in Natural Language Processing}. Association for Computational Linguistics, 2017.

\bibitem{xie2024efficient}
Yong Xie, Karan Aggarwal, and Aitzaz Ahmad.
\newblock Efficient continual pre-training for building domain specific large language models.
\newblock In {\em Findings of the Association for Computational Linguistics ACL 2024}, pages 10184--10201, 2024.

\bibitem{xie2023data}
Sang~Michael Xie, Shibani Santurkar, Tengyu Ma, and Percy~S Liang.
\newblock Data selection for language models via importance resampling.
\newblock {\em Advances in Neural Information Processing Systems}, 36:34201--34227, 2023.

\bibitem{grangier2025taskadaptive}
David Grangier, Simin Fan, Skyler Seto, and Pierre Ablin.
\newblock Task-adaptive pretrained language models via clustered-importance sampling.
\newblock In {\em The Thirteenth International Conference on Learning Representations}, 2025.

\bibitem{kang2024get}
Feiyang Kang, Hoang~Anh Just, Yifan Sun, Himanshu Jahagirdar, Yuanzhi Zhang, Rongxing Du, Anit~Kumar Sahu, and Ruoxi Jia.
\newblock Get more for less: Principled data selection for warming up fine-tuning in {LLM}s.
\newblock In {\em The Twelfth International Conference on Learning Representations}, 2024.

\bibitem{yang2024smalltolarge}
Yu~Yang, Siddhartha Mishra, Jeffrey~N Chiang, and Baharan Mirzasoleiman.
\newblock Smalltolarge (s2l): Scalable data selection for fine-tuning large language models by summarizing training trajectories of small models.
\newblock In {\em The Thirty-eighth Annual Conference on Neural Information Processing Systems}, 2024.

\bibitem{xia2024less}
Mengzhou Xia, Sadhika Malladi, Suchin Gururangan, Sanjeev Arora, and Danqi Chen.
\newblock {LESS}: Selecting influential data for targeted instruction tuning.
\newblock In {\em Forty-first International Conference on Machine Learning}, 2024.

\bibitem{abbas2023semdedup}
Amro Abbas, Kushal Tirumala, D{\'a}niel Simig, Surya Ganguli, and Ari~S Morcos.
\newblock Semdedup: Data-efficient learning at web-scale through semantic deduplication.
\newblock {\em arXiv preprint arXiv:2303.09540}, 2023.

\bibitem{yang2023decoding}
Yu~Yang, Aaditya~K Singh, Mostafa Elhoushi, Anas Mahmoud, Kushal Tirumala, Fabian Gloeckle, Baptiste Rozi{\`e}re, Carole-Jean Wu, Ari~S Morcos, and Newsha Ardalani.
\newblock Decoding data quality via synthetic corruptions: Embedding-guided pruning of code data.
\newblock {\em arXiv preprint arXiv:2312.02418}, 2023.

\bibitem{singh2024brevity}
Aaditya~K Singh, Yu~Yang, Kushal Tirumala, Mostafa Elhoushi, and Ari~S Morcos.
\newblock Brevity is the soul of wit: Pruning long files for code generation.
\newblock {\em arXiv preprint arXiv:2407.00434}, 2024.

\bibitem{shum2025predictive}
Kashun Shum, Yuzhen Huang, Hongjian Zou, Ding Qi, Yixuan Liao, Xiaoxin Chen, Qian Liu, and Junxian He.
\newblock Predictive data selection: The data that predicts is the data that teaches.
\newblock {\em arXiv preprint arXiv:2503.00808}, 2025.

\bibitem{lin2021truthfulqa}
Stephanie Lin, Jacob Hilton, and Owain Evans.
\newblock Truthfulqa: Measuring how models mimic human falsehoods, 2021.

\bibitem{hoffmann2022training}
Jordan Hoffmann, Sebastian Borgeaud, Arthur Mensch, Elena Buchatskaya, Trevor Cai, Eliza Rutherford, Diego de~Las Casas, Lisa~Anne Hendricks, Johannes Welbl, Aidan Clark, et~al.
\newblock Training compute-optimal large language models.
\newblock {\em arXiv preprint arXiv:2203.15556}, 2022.

\bibitem{muennighoff2023scaling}
Niklas Muennighoff, Alexander Rush, Boaz Barak, Teven Le~Scao, Nouamane Tazi, Aleksandra Piktus, Sampo Pyysalo, Thomas Wolf, and Colin~A Raffel.
\newblock Scaling data-constrained language models.
\newblock {\em Advances in Neural Information Processing Systems}, 36:50358--50376, 2023.

\bibitem{goyal2024scaling}
Sachin Goyal, Pratyush Maini, Zachary~C Lipton, Aditi Raghunathan, and J~Zico Kolter.
\newblock Scaling laws for data filtering--data curation cannot be compute agnostic.
\newblock In {\em Proceedings of the IEEE/CVF Conference on Computer Vision and Pattern Recognition}, pages 22702--22711, 2024.

\bibitem{hurst2024gpt}
Aaron Hurst, Adam Lerer, Adam~P Goucher, Adam Perelman, Aditya Ramesh, Aidan Clark, AJ~Ostrow, Akila Welihinda, Alan Hayes, Alec Radford, et~al.
\newblock Gpt-4o system card.
\newblock {\em arXiv preprint arXiv:2410.21276}, 2024.

\end{thebibliography}
\bibliographystyle{unsrt}


\appendix


\newpage

\section{Limitations}
\label{appendix:limitations}
While our proposed {\ModelName} framework demonstrates strong performance and provides valuable insights into data mixture optimization, several limitations warrant further exploration.

First, although we mitigate computational overhead by leveraging lightweight proxy models during the iterative search process, training these proxy models still incurs non-negligible costs. 
We show that using lightweight models (e.g., 350M and 62M) is feasible, but further reducing the computational burden—perhaps through parameter-efficient tuning, distillation, or zero-shot evaluation strategies—remains an important direction for future work.

Second, our evaluation of domain-specific benefits is based on MMLU’s coarse-grained domain categories (e.g., STEM, Social Sciences), which may not fully reflect real-world applications. 
While our results highlight the potential of {\ModelName} for domain adaptation, we have not yet evaluated its effectiveness in high-stakes domains such as finance or healthcare, where data characteristics and requirements can differ substantially. 
We leave such real-world validation to future research.

\section{Societal Impacts}
\label{appendix:societal_impacts}
This work contributes to advancing the field of machine learning by proposing an automated framework for optimizing data mixtures in language model pre-training. 
Our method enables more efficient and scalable objective-aware pre-training by leveraging clustering and iterative search, reducing reliance on manually curated datasets.

From an ethical standpoint, our approach does not introduce new risks beyond those commonly associated with LLM training, such as biases in training data and potential misuse. 
However, optimizing the data mixture may lead to an overrepresentation of certain domains while underrepresenting others, potentially reinforcing knowledge disparities. 
While our method allows for more controlled and targeted data composition, future work should explore safeguards to ensure fairness and mitigate unintended biases in domain representation. 
Additionally, our method automates domain-aware data selection without requiring explicit human annotation, potentially reducing reliance on curated datasets. 
However, this also raises concerns about unintended biases in automatically clustered data.
Ensuring transparency and interpretability in data selection remains a critical area for future exploration.

\section{Experimental Settings}
\label{appendix:experimental_settings}

\subsection{Baselines}
\label{appendix:baselines}
We compare our method with state-of-the-art data mixture methods, including DoReMi~\citep{xie2024doremi}, and RegMix~\citep{liu2024regmix}. 

\begin{itemize}
[leftmargin=*,label=$\bullet$,noitemsep,partopsep=0pt,topsep=0pt,parsep=0pt]
    \item Random: randomly select data for language model training, where each cluster is assigned an equal and uniform weight.
    \item DoReMi~\citep{xie2024doremi}:  a method that trains a small proxy model with group distributionally robust optimization (Group DRO) to determine domain weights for pre-training data, which are then used to resample the dataset and train a larger model more efficiently.
    \item RegMix~\citep{liu2024regmix}: an approach that performs a single round of data mixture configuration search by sampling configurations and training the model on each configuration to obtain config-performance pairs. 
    We then train a regression model to predict the optimal data mixture weights. 
    This method can be regarded as an extension of RegMix~\citep{liu2024regmix}, using a much larger proxy model and larger cluster data labeled by our clustering approach.
\end{itemize}

\subsection{Data}
For the source data, we employ Nemotron-CC~\citep{su2024nemotron} -- a large dataset filtered from Common Crawl.
It divides all the data into 20 buckets based on data quality annotation, and we use the subset from the highest-quality bucket.
The hierarchical clustering of this subset results in approximately 800 billion tokens distributed across 21 clusters.

For the downstream evaluation tasks, we conduct experiments on general reasoning benchmarks including PIQA~\citep{bisk2020piqa}, ARC\_C~\citep{allenai:arc}, ARC\_E~\citep{allenai:arc}, HellaSwag~\citep{zellers2019hellaswag}, WinoGrande~\citep{sakaguchi2021winogrande}, TruthfulQA~\citep{lin2021truthfulqa}, and SIQA~\citep{sap2019socialiqa}. 
In this setting, we optimize the model using the validation data of specific tasks and evaluate it on test data from different tasks. 
For optimization, we use the validation data of only PIQA, ARC\_E, and HellaSwag and evaluate the model on the test sets of all these datasets.
We use LM-Evaluation harness~\citep{eval-harness} for evaluation.
Following the setup in~\citep{allal2024SmolLM,smollmevalsetup}, except for MMLU, which is evaluated using a 5-shot setting, all other datasets are evaluated using a 0-shot setting.

\subsection{Model}
Firstly, we conduct phase-1 pre-training to provide a good foundation for all of the following experiments. 
We train three sizes (62M, 350M, and 1B) of standard Transformer decoder-only models with the next-token language modeling loss.
All of them are trained on 10 trillion tokens, similar to~\citep{yang2024qwen2technicalreport} that trained for 12T tokens.
We acknowledge that this over-training does not strictly align with scaling laws~\citep{hoffmann2022training, muennighoff2023scaling, goyal2024scaling}. 
However, since it does not hurt performance, we chose to train on the same amount of data. 
This practice has also been adopted in some recent models; for example, Qwen-2~\citep{yang2024qwen2technicalreport} utilized 12 Trillion tokens of data to train their 500M model.
We use the warmup-stable-decay (WSD) learning rate schedule~\citep{hu2024minicpm} because it supports resuming at any time of the stable stage and we could focus on the data mixing research in the decay stage. 
For the proxy model, we choose 62M and 350M to conduct experiments, as these sizes are computationally efficient for exploring data mixture configurations. 
For the target model, we conduct experiments on all three sizes (62M, 350M, and 1B) to comprehensively evaluate the impact of our approach across different model scales.
After we identify the optimal data mixture, we continue to train the target model on 40B tokens using this new mixture and then compare its performance. Unless otherwise noted, all reported results are obtained from this 40B continuous pre-training.

\subsection{Training Settings}
\label{appendix:training_settings}
For pre-training, we  use AdamW optimizer and 
 set the learning rate to 5e-5 for the stable stage and anneal it to 1e-5.
We use a batch size of 2M tokens throughout the training process, utilizing 256 NVIDIA H100 GPUs.
The training time of a single lightweight proxy model is approximately 45 GPU hours, while the training time of the large target model is around 6,400 GPU hours. 

\newpage

\section{Analysis}
\label{appendix:analysis}
\subsection{Topics of Clusters}
To gain a deeper understanding of the topics covered in each cluster, we conducted an analysis by extracting the topics with GPT-4o~\citep{hurst2024gpt}. 
Specifically, we randomly sampled 100 documents from each cluster and employed GPT-4o to summarize the most representative topics within them. 
The model was instructed to identify the four to seven most relevant topics for each cluster, ensuring a concise yet comprehensive characterization. 
We also recognize that this approach can only provide auxiliary explanations; our goal is to facilitate the understanding of the internal structure of each cluster rather than to make definitive conclusions through topic analysis.

\begin{table}[ht]
    \centering
    \footnotesize
    \caption{Topics of clusters.}
    \begin{tabular}{lc}
    \toprule
    Cluster-ID & Topics  \\ 
    \midrule
    1                & Environment, Public Health, Policy Development, Medical Innovation        \\ 
    2            & Technology, Neurophysiology, Health and Safety, Innovative Research,  Rehabilitation     \\ 
    3              & Restoration Efforts, Climate and Ecosystem, Community Engagement        \\ 
    4               & Diagnostics, Diseases, Prevention and Control           \\ 
    5              & Vehicles, Ecology, Community, Conservation Efforts    \\ 
    6      & Energy, Science, Materials, Nanostrctures, Quantum Computing          \\ 
    7              & Physics, Accelerators, Materials, Architecture, System    \\ 
    8           & Biology, Genetics, Astronomy, Climate Science     \\ 
    9       & Earth Sciences, Space Science, Scientific Collaboration      \\ 
    10        & Health, Symptoms, Treatment, Therapy, Disorders, Conditions     \\ 
    11          & Communication, Biography, History, Society, Policy        \\ 
    12        & Culture, Education, Sustainability, Community, Public Health, Crime, Economy            \\ 
    13         & Arts, Literature, Education, History          \\ 
    14       & Geography, Government, Organization, Religion, Agriculture, Economy, Civilizations           \\ 
    15         & Science, Technology, Education, Engineering, Collaboration       \\ 
    16         & Science, Health, Minerals, Population, Agriculture, Vaccination, Welfare, Management        \\ 
    17          & Role-Playing, Problem Solving, Mathematics, Algorithms          \\ 
    18          & Revolution, Parliament, Efficiency, Communication, Animal Behavior         \\ 
    19           & History, Culture, Economy, Energy, Market, Policy        \\ 
    20          & Python, Code           \\ 
    21           & Government, Law, Scientific Revolution, Music, Literature        \\ 
    \bottomrule
    \end{tabular}
    \label{tab:cluster_topics}
\end{table}

\subsection{Relationship between Clusters and Downstream Tasks}
\label{appendix:relationship}
\begin{figure}[h]
    \centering
    \includegraphics[scale=0.036, trim=0 0 0 0]{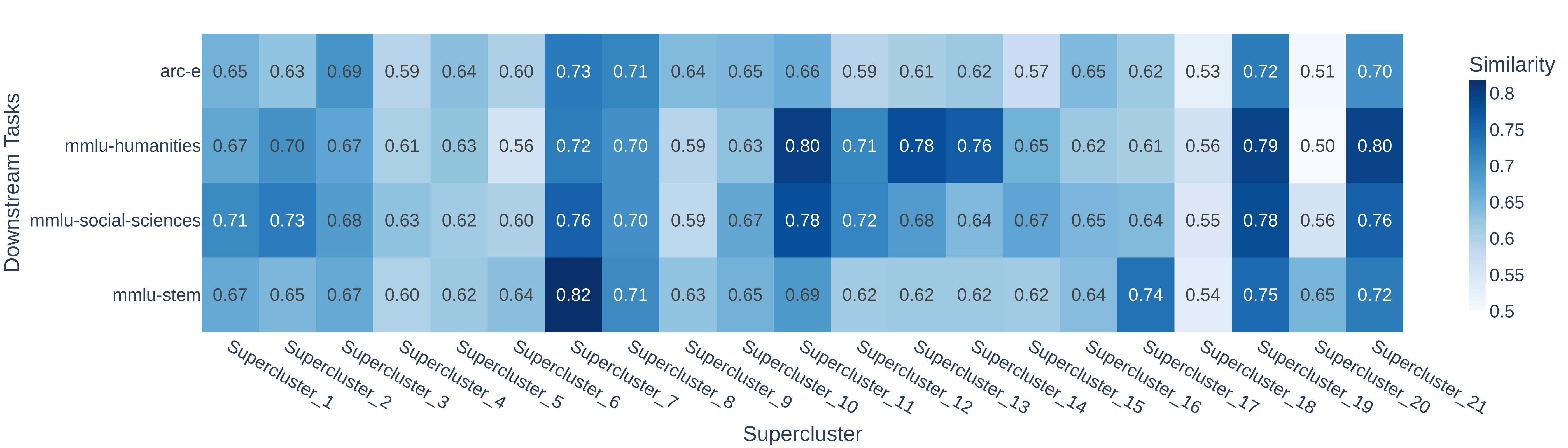}
    \caption{Similarity between clusters and downstream tasks.}
    \label{fig:heatmap-similarity}
\end{figure}

\begin{figure}[H]
    \centering
    \includegraphics[scale=0.26, trim=0 20 0 0]{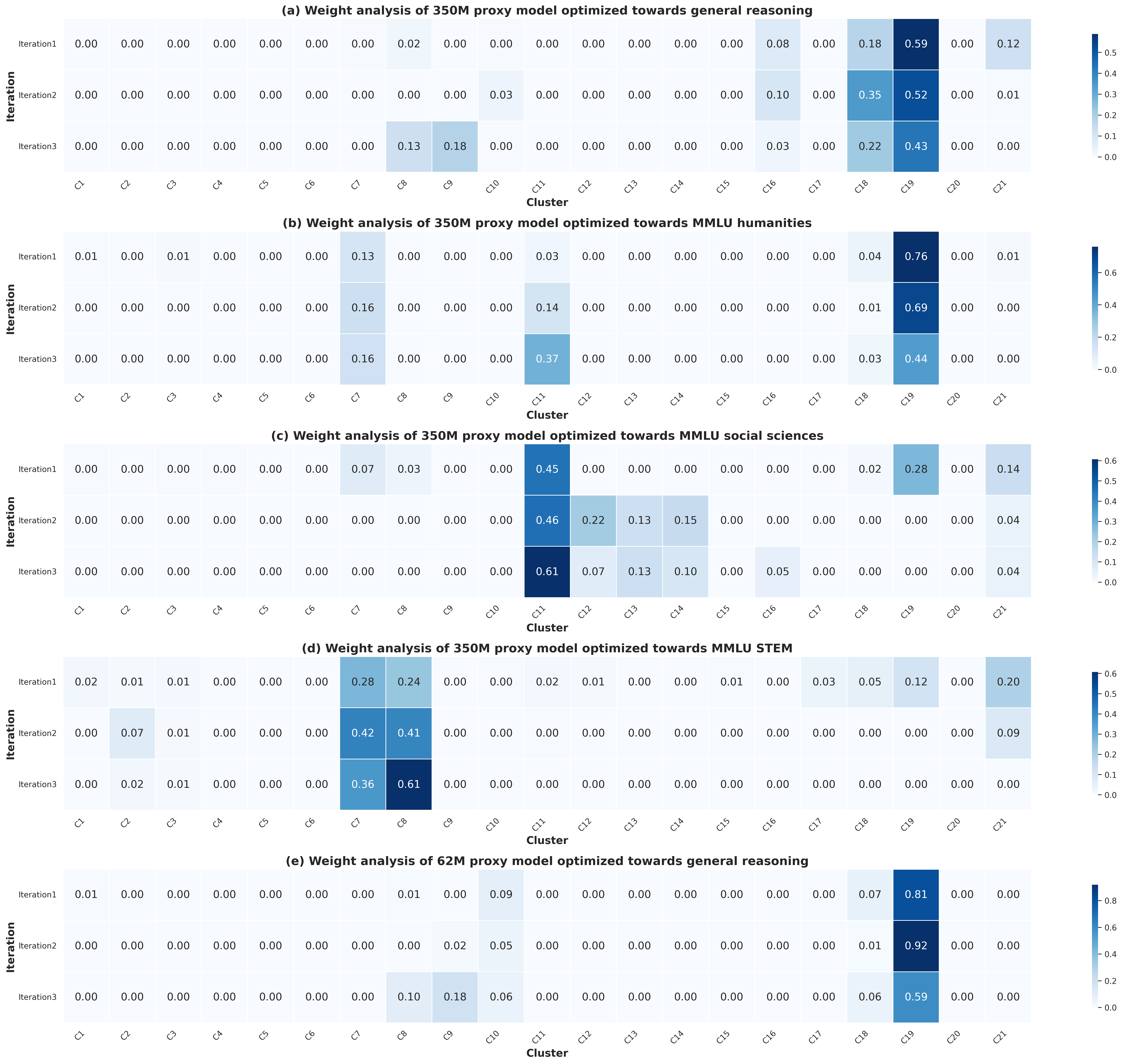}
    \caption{Heatmap of weights across iterations.}
    \label{fig:heatmap-combined}
    \vspace{-1 em}
\end{figure}

In this section, we analyzed the relationship between clusters and downstream task performance.
First, we visualized the similarity between each cluster and downstream tasks in Figure~\ref{fig:heatmap-similarity}, where cosine similarity is measured using the average embedding of each cluster. 
We use arc-e to represent the general reasoning domain. 
Our key observations are as follows: 
(1) in-domain data enhances downstream performance. Take the general reasoning as an example, as shown in Figure~\ref{fig:heatmap-similarity}, Clusters C8 and C19 share the most similar distribution with arc-e and indeed they contribute a lot to the final mixture weights.
(2) out-of-domain data are also useful.
From the results of general reasoning, as our search process iterates, we find that while C21 is highly similar, it provides limited benefits to downstream performance, leading to a gradual decrease in its importance. 
Conversely, C8 initially appears out-of-domain but becomes increasingly important with further iterations.
(3) domain contribution is complex: While similarity can serve as an indicator of a cluster’s importance, it is not always a decisive factor. 
For instance, in mmlu-stem, the most similar cluster is C7, and as shown in Figure~\ref{fig:heatmap-combined} (d), it plays a crucial role, contributing 36\% of the weight. 
However, C8, despite having a lower similarity score, has an even higher weight contribution (61\%).
From this analysis, we observe that highly similar data can sometimes enhance downstream task performance. However, using only in-domain data does not necessarily lead to optimal performance.
Distribution similarity alone is not a sufficient condition for importance—that is, a cluster or domain being similar to a downstream task does not inherently guarantee performance improvement. This is because data mixture involves complex interactions among different clusters. 
In some cases, when clusters are highly similar, incorporating only one of them may suffice. 
This highlights the intricate interplay within data mixtures, suggesting that optimal selection requires more than just similarity-based filtering.

\subsection{Effects of Proxy Model.}
Our method relies on a proxy model to rapidly score candidate mixtures.
Intuitively, larger proxy models should better approximate the performance of the final (larger) target model.
We test three proxy sizes: 62M, 132M, and 350M parameters.
From Table~\ref{tab:ablation-study} (rows under “Abl.proxy”), as we increase the proxy model from 62M to 350M, the average score improves from 60.11 to 60.41.
Although the gains are not dramatic, they consistently favor using the largest feasible proxy model.
This shows that a stronger proxy—closer in capacity to the target—achieves more accurate gradient estimates of mixture quality.

\subsection{Effects of Number of Clusters.}
In our method, we employ a hierarchical clustering procedure. 
Specifically, we first group all data into $K_{init}$ clusters, perform a filtering step, and then regroup these clusters into $K_{enhanced}$ super-clusters. 
In this section, we explore the robustness of our data‐mixture method and investigate its sensitivity to the number of clusters. 
Hence, we experiment with different values of $K_{init}$ (48, 64, 100, 1000, 2000) and $K_{enhanced}$ (15, 21, 30). 
The results in Table~\ref{tab:ablation-study} (rows under “Abl.clus”) show that performance improves as $K_{init}$ increases from 48 to 100 and declines when $K_{init}$ increases from 1000 to 2000.
Overall, our method is not particularly sensitive to the number of clusters, demonstrating the robustness of our approach.
It is worth mentioning that if $K_{init}$ exceeds 2000 given the dataset size, the clustering becomes overly fine‐grained and thus too dispersed. 
Likewise, if $K_{enhanced}$ is set too high, it requires more compute for sampling, increasing the overall cost of the data search process.

\subsection{Effects of Initialization.}
We compare how different initialization schemes for the mixture weights affect performance.
We experiment with a simple random initialization versus a Dirichlet‐based initialization that biases weights to be more evenly spread at the start.
Table~\ref{tab:ablation-study} (rows under “Abl.init”) shows that Dirichlet initialization achieves a slightly higher average score (60.41\%) than random (60.21\%).
The performances are comparable, suggesting the robustness of our data mixing approach, which is largely insensitive to the choice of initialization.

\subsection{Effects of Compute Allocation.}
By default, we allocate our 100\% total compute across three iterations in a 4{:}2{:}1 ratio (64{:}32{:}16).
In principle, however, one could allocate compute to create either a “tall” search tree (more iterations but fewer searches per iteration) or a “fat” one (fewer iterations but more searches per iteration).
Table~\ref{tab:ablation-study} (rows under “Abl.allo”) compares several such allocations: 6{:}1, 4{:}2{:}1, and 2{:}2{:}1{:}1.
We find that 4{:}2{:}1 yields the best overall average performance (60.41\%).
Having too few iterations (e.g., 6{:}1) can lead to suboptimal exploration in earlier iterations, while splitting too many iterations (2{:}2{:}1{:}1) spreads compute too thin across each iteration.
Thus, balancing depth (number of iterations) and breadth (searches per iteration) proves key to robustly finding a good mixture.

\subsection{Evolution of Cluster Weights}
\label{appendix:evolution}
The data mixture weights are important to understand the impact of different clusters, so we closely examine how they evolve across iterations.
Figure~\ref{fig:heatmap-combined} (a) presents the weights discovered by our search process for the 350M proxy model in the general reasoning domain. 
As shown, most clusters have minimal or no contribution (weights close to 0.00), while a few clusters play a significant role, with their weights changing across iterations.
Among them, C18, C19, and C21 initially have high weights, but C19 and C21 exhibit a decreasing trend, suggesting their diminishing impact.
Conversely, C8 and C9 become more relevant in later iterations, with their weights increasing in Iteration 3 (C8: 0.13, C9: 0.18), highlighting an adaptation in feature importance.

\subsection{Analysis of Final Weights}
\label{appendix:analysis_of_final_weights}

Furthermore, we analyzed the weights of the final data mixtures.
From Figure~\ref{fig:heatmap-combined} (a) , for the general reasoning task,
C8, C9, C18, and C19 account for the majority of the weight.
As shown in \ref{appendix:relationship}, C8, C9, and C19 exhibit a high degree of correlation with general reasoning.
Moreover, when analyzing the topics of these four clusters (\ref{tab:cluster_topics}), we find that they collectively form a diverse distribution.

In addition, we analyzed the importance of different clusters across domains on MMLU. 
As shown in Figures~\ref{fig:heatmap-combined} (b), (c), and (d), certain clusters play a crucial role in specific domains. 
For example, C7, C11, and C19 are particularly important for the humanities domain, while C7 and C8 are highly influential in the STEM domain. 
These findings highlight how different clusters contribute uniquely to various domains, providing deeper insights into domain-specific feature significance.
We are also curious about the similarities and differences in the weights discovered by the large proxy model and the small proxy model. 
To explore this, we compared Figure~\ref{fig:heatmap-combined} (a) and (e), and observed that they share similar important features, such as C8, C9, C18, and C19, although the assigned weights vary between the models. 
This insight suggests that we can leverage a smaller 62M proxy model for further experiments, reducing computational costs while retaining key structural patterns. 
The experimental results are presented in Appendix~\ref{appendix:62m-proxy}.
Notably, the weights appear sparse because, during the sampling process, we intentionally bias towards sparse weights. 
This approach effectively amplifies important clusters while filtering out less significant ones, enhancing the clarity of key features.
In addition, we also investigate the relationship between clusters and downstream task performance in \ref{appendix:relationship}.

\subsection{Experiments with 62M proxy model}
\label{appendix:62m-proxy}
In the main experiment, we used a 350M proxy model. 
To further investigate the effectiveness of smaller proxy models, we conducted additional experiments with reduced model sizes. 
The results, presented in Tables~\ref{tab:62m-proxy-mmlu} and ~\ref{tab:62m-proxy-general-reasoning}
, indicate that even when the proxy model size was reduced by a factor of five, its performance remained strong. 
This suggests that smaller proxy models can still be highly effective, providing valuable insights while reducing computational costs.

\begin{table}[H]
\centering
\small
\caption{Performance of target models on MMLU-social-sciences task. 
The main proxy model is 62M.}
\begin{tabular}{c|cc|c}
\toprule
\textbf{Target} & \textbf{Model}  & \textbf{Proxy} & \textbf{Accuracy (\%)} \\ 
\midrule
\multirow{6}{*}{62M} 
 & Random  & - & 27.40 \\ 
 & {\ModelName}-Best@N  & 62M & 31.03 \\ 
 & {\ModelName}-iter1  & 62M & 29.05 \\ 
 & {\ModelName}-iter2 & 62M & 30.71 \\ 
 & {\ModelName}-iter3  & 62M & 32.43 \\ 
\midrule
\multirow{6}{*}{350M} 
 & Random  & - & 34.87 \\ 
 & {\ModelName}-Best@N  & 350M & 38.39 \\ 
 & {\ModelName}-iter1  & 62M & 36.09 \\ 
 & {\ModelName}-iter2  & 62M & 37.01 \\ 
 & {\ModelName}-iter3  & 62M & 37.98 \\ 
\midrule
\multirow{6}{*}{1B} 
 & Random  & - & 36.69 \\ 
 & {\ModelName}-Best@N & 1B & 40.66 \\ 
 & {\ModelName}-iter1  & 62M & 40.03 \\ 
 & {\ModelName}-iter2  & 62M & 40.46 \\ 
 & {\ModelName}-iter3  & 62M & 41.72 \\ 
\bottomrule
\end{tabular}%
\label{tab:62m-proxy-mmlu}
\end{table}

\begin{table}[H]
\centering
\setlength{\tabcolsep}{4pt}
\caption{Performance of target models on general reasoning benchmarks.
The main proxy model is 62M.}
\resizebox{0.9\textwidth}{!}{%
\begin{tabular}{c|ccc|ccccccc}
\toprule
{\textbf{Size}} & {\textbf{Model}} & {\textbf{Proxy}} & {\textbf{Comp.}} & \textbf{piqa} & \textbf{arc\_c} & \textbf{arc\_e} & \textbf{hellaswag} & \textbf{winogrande} & \textbf{siqa} & \textbf{Avg.} \\ 
\midrule
\multirow{6}{*}{62M} 
 & Random  & - & 0 & 61.80 & 24.06 & 45.70 & 33.64 & 50.19 & 37.51 & 41.76 \\ 
 & {\ModelName}-Best@N  & 62M & 100\% & 63.16 & 25.51 & 51.30 & 35.68 & 51.14 & 38.07 & 44.14 \\
 & {\ModelName}-iter1  & 62M & 57\% & 63.92 & 24.82 & 49.83 & 34.76 & 49.48 & 38.79 & 43.60  \\ 
 & {\ModelName}-iter2  & 62M & 85\% & 64.09 & 26.10 & 49.83 & 35.95 & 51.06 & 38.68 & 44.29 \\ 
 & {\ModelName}-iter3 & 62M & 100\% & 64.54 & 27.01 & 53.39 & 35.82 & 51.15 & 39.50 & 45.23 \\  
\midrule
\multirow{6}{*}{350M} 
 & Random  & - & 0 & 71.16 & 30.54 & 62.50 & 52.14 & 55.40 & 41.29 & 52.17 \\ 
 & {\ModelName}-Best@N  & 350M & 100\% & 71.92 & 33.70 & 67.00 & 54.55 & 56.59 & 41.67 & 54.24 \\ 
 & {\ModelName}-iter1  & 62M & 57\%  & 71.65 & 33.49 & 65.31 & 54.44 & 56.28 & 41.99 & 53.86 \\ 
 & {\ModelName}-iter2 & 62M & 85\%  & 71.54 & 34.01 & 66.43 & 54.61 & 56.78 & 41.37 & 54.12 \\ 
 & {\ModelName}-iter3  & 62M & 100\% & 71.87 & 34.12 & 66.92 & 54.81 & 56.11 & 42.37 & 54.37 \\
\midrule
\multirow{6}{*}{1B} 
 & Random  & - & 0 & 74.05 & 37.12 & 70.24 & 62.90 & 60.77 & 42.48 & 57.93 \\ 
 & {\ModelName}-Best@N  & 1B & 100\% & 75.02 & 38.39 & 72.34 & 64.31 & 61.16 & 42.52 & 58.96 \\ 
 & {\ModelName}-iter1 & 62M & 57\% & 74.38 & 38.19 & 70.98 & 64.21 & 61.58 & 43.11 & 58.74 \\ 
 & {\ModelName}-iter2 & 62M & 85\% & 75.26 & 39.28 & 72.17 & 63.99 & 63.16 & 41.27 & 59.19 \\ 
 & {\ModelName}-iter3  & 62M & 100\% & 75.41 & 40.56 & 72.82 & 65.76 & 63.23 & 42.89 & 60.11 \\
\bottomrule
\end{tabular}%
}
\label{tab:62m-proxy-general-reasoning}
\end{table}

\begin{figure}[t]
    \centering
    \includegraphics[scale=0.52, trim=0 10 0 0]{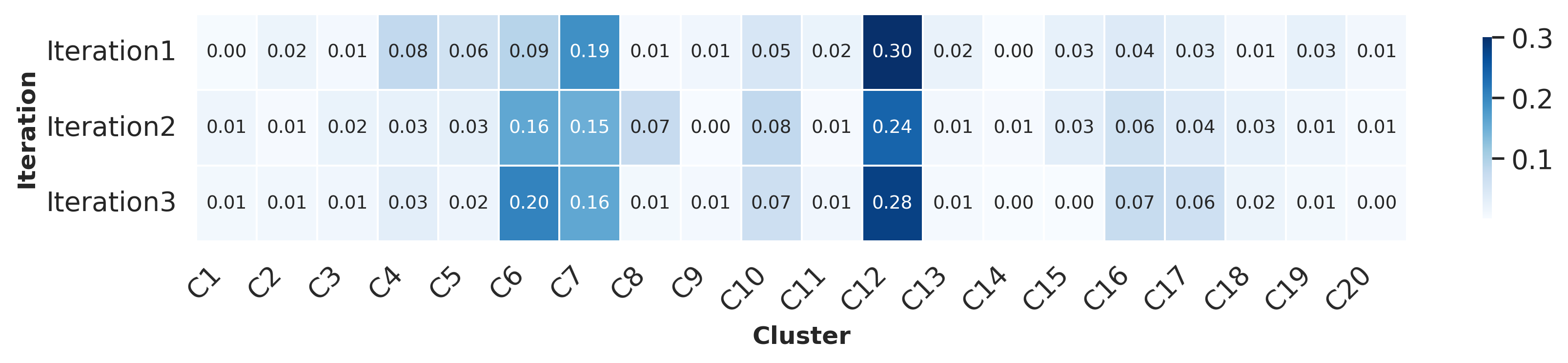}
    \caption{Weight analysis of {\finaldatamix} across iterations.}
    \label{fig:heatmap-clibmix}
    \vspace{-2 em}
\end{figure}

\subsection{Effects of Predictor}
\begin{figure}[H]
    \centering
    \includegraphics[scale=0.4, trim=0 0 0 0]{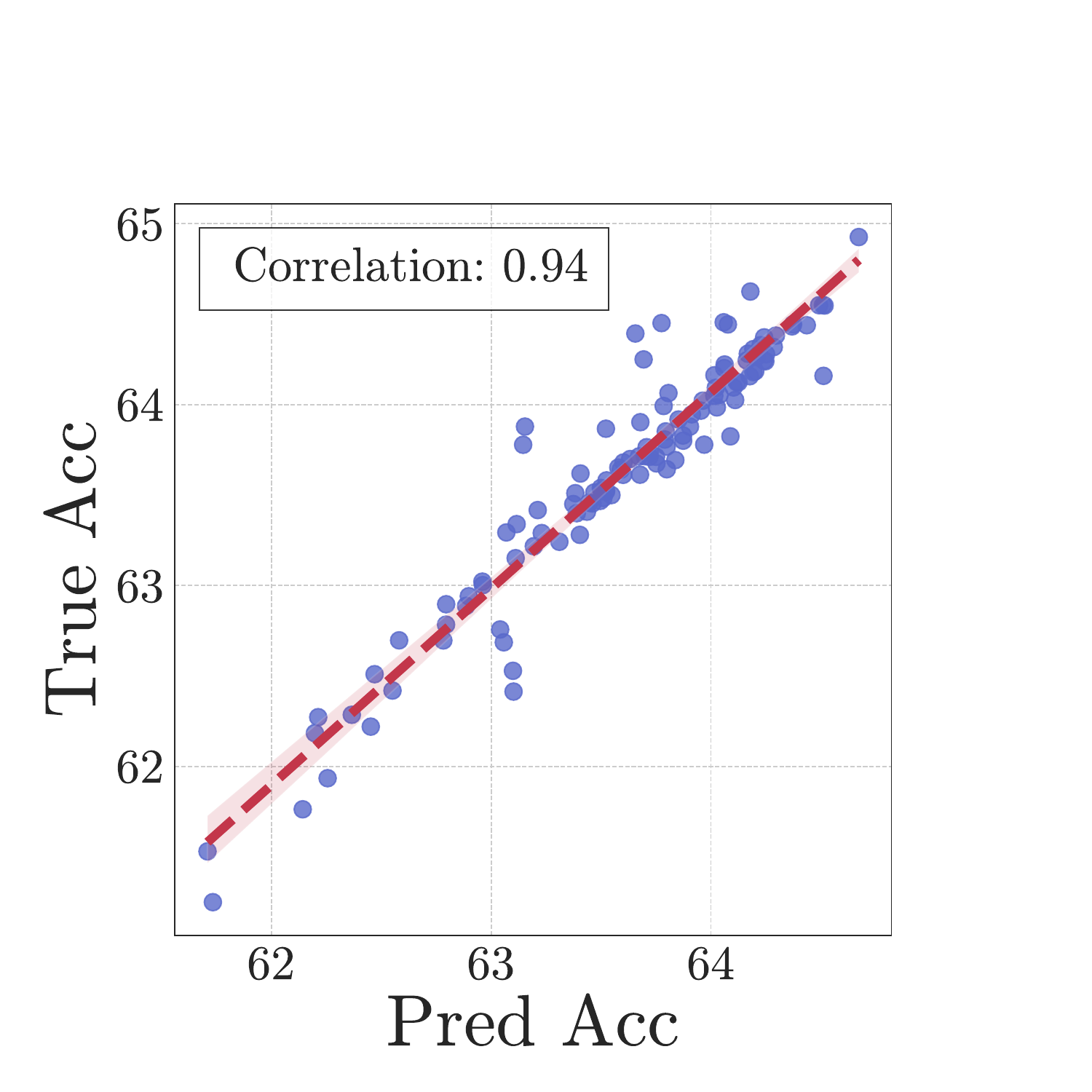}
    \caption{The Spearman rank correlation between predicted accuracy made by the predictor model and the groundtruth accuracy.}
    \label{fig:predictor}
    \vspace{-1 em}
\end{figure}

In our approach, after training the proxy model on configuration-performance pairs, we use a regression model (i.e., predictor) to capture the relationship between configuration and target domain performance. 
To evaluate prediction accuracy, we hold out a portion of the data as the test set and compute the Spearman rank correlation between the predictions and ground truth. 
As shown in Figure~\ref{fig:predictor}, we visualize the predicted and true accuracy pairs for the 350M proxy models and find that the predictor performs exceptionally well, achieving  94\% Spearman rank correlation.

\subsection{Prompt Template}
\label{appendix:prompt}
We present the prompts used for data annotation, as shown in the table below.

\begin{tcolorbox}[
    float,
    colback=lightgray, colframe=darkblue, 
    title=Evaluation Criteria for Pre-Training Data,
    fonttitle=\bfseries, sharp corners, boxrule=1pt
]
You are an expert evaluator assessing a text for suitability as pre-training data for a large language model. 
For each criterion, start from 0 points. Then add points based on the conditions described. If no conditions are met, the score remains 0.
Please evaluate the given text using the rating scale below. 
Assign a score from 0 to 5 for each criterion, and reference the expanded guidelines under each category to determine the appropriate rating:

\textbf{Rating Scale:}
\vspace{-5pt}
\begin{itemize}
    \setlength{\itemsep}{0pt}
    \setlength{\parskip}{0pt}
    \item \textbf{0}: Does not meet the criterion at all
    \item \textbf{1}: Partially meets the criterion
    \item \textbf{2}: Fairly meets the criterion
    \item \textbf{3}: Mostly meets the criterion
    \item \textbf{4}: Fully meets the criterion
    \item \textbf{5}: Exceeds the criterion
\end{itemize}
\vspace{-5pt}

\textbf{Criteria and Expanded Guidelines:}
\vspace{-5pt} 
\begin{enumerate}
    \setlength{\itemsep}{0pt}
    \setlength{\parskip}{0pt}
    \item \textbf{Quality}: The text is natural, clean, and free from severe grammatical errors, spelling mistakes, syntactical issues, repetitive phrasing, or random symbols.
    \begin{itemize}
    \setlength{\itemsep}{0pt}
    \setlength{\parskip}{0pt}
        \item +1: Correct basic spelling and mostly proper grammar, despite minor slips.
        \item +1: Coherent sentence structures, no glaring syntactical breakdowns.
        \item +1: Natural language, free from repetitive phrasing, easy to read.
        \item +1: Polished, no major grammatical errors or spelling mistakes.
        \item +1: Professional-level writing quality, free from unnatural phrasing.
    \end{itemize}
    
    \item \textbf{Advertisement}: The text should avoid excessive promotional language or overt advertising.
    \begin{itemize}
        \setlength{\itemsep}{0pt}
    \setlength{\parskip}{0pt}
        \item +1: Minimal promotional elements, not distracting.
        \item +1: Subtle promotional aspects, not overshadowing content.
        \item +1: Mostly neutral with slight marketing-like language.
        \item +1: Almost free from advertisements, at most one mild reference.
        \item +1: No detectable promotional content.
    \end{itemize}

    \item \textbf{Informational Value}: The text provides accurate insights, useful facts, or relevant knowledge.
    \begin{itemize}
        \setlength{\itemsep}{0pt}
    \setlength{\parskip}{0pt}
        \item +1: At least one accurate fact or relevant information.
        \item +1: Multiple useful pieces of information.
        \item +1: Enhances understanding, presents explanations.
        \item +1: Substantial, well-structured, reliable information.
        \item +1: Exceptional depth, authoritative content.
    \end{itemize}

    \item \textbf{Educational Value}: Assess if the text is beneficial for structured learning.
    \begin{itemize}
        \setlength{\itemsep}{0pt}
    \setlength{\parskip}{0pt}
        \item +1: Basic educational relevance, even if mixed with non-academic content.
        \item +1: Addresses education but lacks strong alignment with standards.
        \item +1: Suitable for educational use, introduces key concepts.
        \item +1: Highly relevant for structured learning, minimal extraneous content.
        \item +1: Outstanding educational value, clear, easy-to-follow insights.
    \end{itemize}
\end{enumerate}

\textbf{Final Output Format:}
\vspace{-5pt}

\begin{lstlisting}[language=Python, frame=single, backgroundcolor=\color{lightgray}, basicstyle=\ttfamily]
{
    "quality": < integer 0-5 >,
    "advertisement": < integer 0-5 >,
    "informational_value": < integer 0-5 >,
    "educational_value": < integer 0-5 >,
}
\end{lstlisting}
\vspace{-5pt}

\textbf{Content to evaluate:}
\vspace{-5pt}
\begin{tcolorbox}[colback=white, colframe=gray, sharp corners]
\texttt{INPUT\_DOC}
\end{tcolorbox}

\end{tcolorbox}

\clearpage
\newpage
\section*{NeurIPS Paper Checklist}

\begin{enumerate}

\item {\bf Claims}
    \item[] Question: Do the main claims made in the abstract and introduction accurately reflect the paper's contributions and scope?
    \item[] Answer: \answerYes{} 
    \item[] Justification: We present extensive quantitative experiments and demonstrate our claims across a wide variety of benchmarks.
    \item[] Guidelines: 
    \begin{itemize}
        \item The answer NA means that the abstract and introduction do not include the claims made in the paper.
        \item The abstract and/or introduction should clearly state the claims made, including the contributions made in the paper and important assumptions and limitations. A No or NA answer to this question will not be perceived well by the reviewers. 
        \item The claims made should match theoretical and experimental results, and reflect how much the results can be expected to generalize to other settings. 
        \item It is fine to include aspirational goals as motivation as long as it is clear that these goals are not attained by the paper. 
    \end{itemize}

\item {\bf Limitations}
    \item[] Question: Does the paper discuss the limitations of the work performed by the authors?
    \item[] Answer: \answerYes{} 
    \item[] Justification: Limitations are discussed in Appendix~\ref{appendix:limitations}.
    \item[] Guidelines:
    \begin{itemize}
        \item The answer NA means that the paper has no limitation while the answer No means that the paper has limitations, but those are not discussed in the paper. 
        \item The authors are encouraged to create a separate "Limitations" section in their paper.
        \item The paper should point out any strong assumptions and how robust the results are to violations of these assumptions (e.g., independence assumptions, noiseless settings, model well-specification, asymptotic approximations only holding locally). The authors should reflect on how these assumptions might be violated in practice and what the implications would be.
        \item The authors should reflect on the scope of the claims made, e.g., if the approach was only tested on a few datasets or with a few runs. In general, empirical results often depend on implicit assumptions, which should be articulated.
        \item The authors should reflect on the factors that influence the performance of the approach. For example, a facial recognition algorithm may perform poorly when image resolution is low or images are taken in low lighting. Or a speech-to-text system might not be used reliably to provide closed captions for online lectures because it fails to handle technical jargon.
        \item The authors should discuss the computational efficiency of the proposed algorithms and how they scale with dataset size.
        \item If applicable, the authors should discuss possible limitations of their approach to address problems of privacy and fairness.
        \item While the authors might fear that complete honesty about limitations might be used by reviewers as grounds for rejection, a worse outcome might be that reviewers discover limitations that aren't acknowledged in the paper. The authors should use their best judgment and recognize that individual actions in favor of transparency play an important role in developing norms that preserve the integrity of the community. Reviewers will be specifically instructed to not penalize honesty concerning limitations.
    \end{itemize}

\item {\bf Theory assumptions and proofs}
    \item[] Question: For each theoretical result, does the paper provide the full set of assumptions and a complete (and correct) proof?
    \item[] Answer: \answerNA{} 
    \item[] Justification: Our paper does not contain theoretical results.
    \item[] Guidelines:
    \begin{itemize}
        \item The answer NA means that the paper does not include theoretical results. 
        \item All the theorems, formulas, and proofs in the paper should be numbered and cross-referenced.
        \item All assumptions should be clearly stated or referenced in the statement of any theorems.
        \item The proofs can either appear in the main paper or the supplemental material, but if they appear in the supplemental material, the authors are encouraged to provide a short proof sketch to provide intuition. 
        \item Inversely, any informal proof provided in the core of the paper should be complemented by formal proofs provided in appendix or supplemental material.
        \item Theorems and Lemmas that the proof relies upon should be properly referenced. 
    \end{itemize}

    \item {\bf Experimental result reproducibility}
    \item[] Question: Does the paper fully disclose all the information needed to reproduce the main experimental results of the paper to the extent that it affects the main claims and/or conclusions of the paper (regardless of whether the code and data are provided or not)?
    \item[] Answer: \answerYes{} 
    \item[] Justification: We provide detailed descriptions of our framework, training setups, proxy model configurations, clustering procedures, and evaluation metrics in the main paper and appendix. 
    In addition, we release the datasets used in our experiments ({\climblab} and {\finaldatamix}) and the corresponding Hugging Face link is included in the abstract. These resources, together with the methodology described, are sufficient for reproducing the main experimental results and validating the conclusions.
    \item[] Guidelines:
    \begin{itemize}
        \item The answer NA means that the paper does not include experiments.
        \item If the paper includes experiments, a No answer to this question will not be perceived well by the reviewers: Making the paper reproducible is important, regardless of whether the code and data are provided or not.
        \item If the contribution is a dataset and/or model, the authors should describe the steps taken to make their results reproducible or verifiable. 
        \item Depending on the contribution, reproducibility can be accomplished in various ways. For example, if the contribution is a novel architecture, describing the architecture fully might suffice, or if the contribution is a specific model and empirical evaluation, it may be necessary to either make it possible for others to replicate the model with the same dataset, or provide access to the model. In general. releasing code and data is often one good way to accomplish this, but reproducibility can also be provided via detailed instructions for how to replicate the results, access to a hosted model (e.g., in the case of a large language model), releasing of a model checkpoint, or other means that are appropriate to the research performed.
        \item While NeurIPS does not require releasing code, the conference does require all submissions to provide some reasonable avenue for reproducibility, which may depend on the nature of the contribution. For example
        \begin{enumerate}
            \item If the contribution is primarily a new algorithm, the paper should make it clear how to reproduce that algorithm.
            \item If the contribution is primarily a new model architecture, the paper should describe the architecture clearly and fully.
            \item If the contribution is a new model (e.g., a large language model), then there should either be a way to access this model for reproducing the results or a way to reproduce the model (e.g., with an open-source dataset or instructions for how to construct the dataset).
            \item We recognize that reproducibility may be tricky in some cases, in which case authors are welcome to describe the particular way they provide for reproducibility. In the case of closed-source models, it may be that access to the model is limited in some way (e.g., to registered users), but it should be possible for other researchers to have some path to reproducing or verifying the results.
        \end{enumerate}
    \end{itemize}

\item {\bf Open access to data and code}
    \item[] Question: Does the paper provide open access to the data and code, with sufficient instructions to faithfully reproduce the main experimental results, as described in supplemental material?
    \item[] Answer: \answerYes{} 
    \item[] Justification: We release our curated data and training code \href{https://huggingface.co/collections/nvidia/climb-datasets-67e428bdb9aaced2acda191f}{here}.
    \item[] Guidelines:
    \begin{itemize}
        \item The answer NA means that paper does not include experiments requiring code.
        \item Please see the NeurIPS code and data submission guidelines (\url{https://nips.cc/public/guides/CodeSubmissionPolicy}) for more details.
        \item While we encourage the release of code and data, we understand that this might not be possible, so “No” is an acceptable answer. Papers cannot be rejected simply for not including code, unless this is central to the contribution (e.g., for a new open-source benchmark).
        \item The instructions should contain the exact command and environment needed to run to reproduce the results. See the NeurIPS code and data submission guidelines (\url{https://nips.cc/public/guides/CodeSubmissionPolicy}) for more details.
        \item The authors should provide instructions on data access and preparation, including how to access the raw data, preprocessed data, intermediate data, and generated data, etc.
        \item The authors should provide scripts to reproduce all experimental results for the new proposed method and baselines. If only a subset of experiments are reproducible, they should state which ones are omitted from the script and why.
        \item At submission time, to preserve anonymity, the authors should release anonymized versions (if applicable).
        \item Providing as much information as possible in supplemental material (appended to the paper) is recommended, but including URLs to data and code is permitted.
    \end{itemize}

\item {\bf Experimental setting/details}
    \item[] Question: Does the paper specify all the training and test details (e.g., data splits, hyperparameters, how they were chosen, type of optimizer, etc.) necessary to understand the results?
    \item[] Answer: \answerYes{} 
    \item[] Justification: We provide training details in Appendix~\ref{appendix:experimental_settings}.
    \item[] Guidelines:
    \begin{itemize}
        \item The answer NA means that the paper does not include experiments.
        \item The experimental setting should be presented in the core of the paper to a level of detail that is necessary to appreciate the results and make sense of them.
        \item The full details can be provided either with the code, in appendix, or as supplemental material.
    \end{itemize}

\item {\bf Experiment statistical significance}
    \item[] Question: Does the paper report error bars suitably and correctly defined or other appropriate information about the statistical significance of the experiments?
    \item[] Answer: \answerNo{} 
    \item[] Justification: We performed relatively large-scale training ($\geq$ 100B tokens) so that the task performance is stable across runs.
    \item[] Guidelines:
    \begin{itemize}
        \item The answer NA means that the paper does not include experiments.
        \item The authors should answer "Yes" if the results are accompanied by error bars, confidence intervals, or statistical significance tests, at least for the experiments that support the main claims of the paper.
        \item The factors of variability that the error bars are capturing should be clearly stated (for example, train/test split, initialization, random drawing of some parameter, or overall run with given experimental conditions).
        \item The method for calculating the error bars should be explained (closed form formula, call to a library function, bootstrap, etc.)
        \item The assumptions made should be given (e.g., Normally distributed errors).
        \item It should be clear whether the error bar is the standard deviation or the standard error of the mean.
        \item It is OK to report 1-sigma error bars, but one should state it. The authors should preferably report a 2-sigma error bar than state that they have a 96\% CI, if the hypothesis of Normality of errors is not verified.
        \item For asymmetric distributions, the authors should be careful not to show in tables or figures symmetric error bars that would yield results that are out of range (e.g. negative error rates).
        \item If error bars are reported in tables or plots, The authors should explain in the text how they were calculated and reference the corresponding figures or tables in the text.
    \end{itemize}

\item {\bf Experiments compute resources}
    \item[] Question: For each experiment, does the paper provide sufficient information on the computer resources (type of compute workers, memory, time of execution) needed to reproduce the experiments?
    \item[] Answer: \answerYes{} 
    \item[] Justification: The training resources are detailed in Appendix~\ref{appendix:training_settings}.
    \item[] Guidelines:
    \begin{itemize}
        \item The answer NA means that the paper does not include experiments.
        \item The paper should indicate the type of compute workers CPU or GPU, internal cluster, or cloud provider, including relevant memory and storage.
        \item The paper should provide the amount of compute required for each of the individual experimental runs as well as estimate the total compute. 
        \item The paper should disclose whether the full research project required more compute than the experiments reported in the paper (e.g., preliminary or failed experiments that didn't make it into the paper). 
    \end{itemize}
    
\item {\bf Code of ethics}
    \item[] Question: Does the research conducted in the paper conform, in every respect, with the NeurIPS Code of Ethics \url{https://neurips.cc/public/EthicsGuidelines}?
    \item[] Answer: \answerYes{} 
    \item[] Justification: We have read and followed this guideline.
    \item[] Guidelines:
    \begin{itemize}
        \item The answer NA means that the authors have not reviewed the NeurIPS Code of Ethics.
        \item If the authors answer No, they should explain the special circumstances that require a deviation from the Code of Ethics.
        \item The authors should make sure to preserve anonymity (e.g., if there is a special consideration due to laws or regulations in their jurisdiction).
    \end{itemize}

\item {\bf Broader impacts}
    \item[] Question: Does the paper discuss both potential positive societal impacts and negative societal impacts of the work performed?
    \item[] Answer: \answerYes{} 
    \item[] Justification: We discuss it in Appendix~\ref{appendix:societal_impacts}.
    \item[] Guidelines:
    \begin{itemize}
        \item The answer NA means that there is no societal impact of the work performed.
        \item If the authors answer NA or No, they should explain why their work has no societal impact or why the paper does not address societal impact.
        \item Examples of negative societal impacts include potential malicious or unintended uses (e.g., disinformation, generating fake profiles, surveillance), fairness considerations (e.g., deployment of technologies that could make decisions that unfairly impact specific groups), privacy considerations, and security considerations.
        \item The conference expects that many papers will be foundational research and not tied to particular applications, let alone deployments. However, if there is a direct path to any negative applications, the authors should point it out. For example, it is legitimate to point out that an improvement in the quality of generative models could be used to generate deepfakes for disinformation. On the other hand, it is not needed to point out that a generic algorithm for optimizing neural networks could enable people to train models that generate Deepfakes faster.
        \item The authors should consider possible harms that could arise when the technology is being used as intended and functioning correctly, harms that could arise when the technology is being used as intended but gives incorrect results, and harms following from (intentional or unintentional) misuse of the technology.
        \item If there are negative societal impacts, the authors could also discuss possible mitigation strategies (e.g., gated release of models, providing defenses in addition to attacks, mechanisms for monitoring misuse, mechanisms to monitor how a system learns from feedback over time, improving the efficiency and accessibility of ML).
    \end{itemize}
    
\item {\bf Safeguards}
    \item[] Question: Does the paper describe safeguards that have been put in place for responsible release of data or models that have a high risk for misuse (e.g., pretrained language models, image generators, or scraped datasets)?
    \item[] Answer: \answerNo{} 
    \item[] Justification: Our data is based on publicly available open datasets, so we did not apply additional safeguards to further enhance safety.
    \item[] Guidelines:
    \begin{itemize}
        \item The answer NA means that the paper poses no such risks.
        \item Released models that have a high risk for misuse or dual-use should be released with necessary safeguards to allow for controlled use of the model, for example by requiring that users adhere to usage guidelines or restrictions to access the model or implementing safety filters. 
        \item Datasets that have been scraped from the Internet could pose safety risks. The authors should describe how they avoided releasing unsafe images.
        \item We recognize that providing effective safeguards is challenging, and many papers do not require this, but we encourage authors to take this into account and make a best faith effort.
    \end{itemize}

\item {\bf Licenses for existing assets}
    \item[] Question: Are the creators or original owners of assets (e.g., code, data, models), used in the paper, properly credited and are the license and terms of use explicitly mentioned and properly respected?
    \item[] Answer: \answerYes{} 
    \item[] Justification: We used open data and have provided the corresponding references.
    \item[] Guidelines:
    \begin{itemize}
        \item The answer NA means that the paper does not use existing assets.
        \item The authors should cite the original paper that produced the code package or dataset.
        \item The authors should state which version of the asset is used and, if possible, include a URL.
        \item The name of the license (e.g., CC-BY 4.0) should be included for each asset.
        \item For scraped data from a particular source (e.g., website), the copyright and terms of service of that source should be provided.
        \item If assets are released, the license, copyright information, and terms of use in the package should be provided. For popular datasets, \url{paperswithcode.com/datasets} has curated licenses for some datasets. Their licensing guide can help determine the license of a dataset.
        \item For existing datasets that are re-packaged, both the original license and the license of the derived asset (if it has changed) should be provided.
        \item If this information is not available online, the authors are encouraged to reach out to the asset's creators.
    \end{itemize}

\item {\bf New assets}
    \item[] Question: Are new assets introduced in the paper well documented and is the documentation provided alongside the assets?
    \item[] Answer: \answerYes{} 
    \item[] Justification: Yes. The documentation is provided in the dataset card hosted on Hugging Face.
    \item[] Guidelines:
    \begin{itemize}
        \item The answer NA means that the paper does not release new assets.
        \item Researchers should communicate the details of the dataset/code/model as part of their submissions via structured templates. This includes details about training, license, limitations, etc. 
        \item The paper should discuss whether and how consent was obtained from people whose asset is used.
        \item At submission time, remember to anonymize your assets (if applicable). You can either create an anonymized URL or include an anonymized zip file.
    \end{itemize}

\item {\bf Crowdsourcing and research with human subjects}
    \item[] Question: For crowdsourcing experiments and research with human subjects, does the paper include the full text of instructions given to participants and screenshots, if applicable, as well as details about compensation (if any)? 
    \item[] Answer: \answerNo{} 
    \item[] Justification: No crowdsourcing experiments or research involving human subjects were conducted.
    \item[] Guidelines:
    \begin{itemize}
        \item The answer NA means that the paper does not involve crowdsourcing nor research with human subjects.
        \item Including this information in the supplemental material is fine, but if the main contribution of the paper involves human subjects, then as much detail as possible should be included in the main paper. 
        \item According to the NeurIPS Code of Ethics, workers involved in data collection, curation, or other labor should be paid at least the minimum wage in the country of the data collector. 
    \end{itemize}

\item {\bf Institutional review board (IRB) approvals or equivalent for research with human subjects}
    \item[] Question: Does the paper describe potential risks incurred by study participants, whether such risks were disclosed to the subjects, and whether Institutional Review Board (IRB) approvals (or an equivalent approval/review based on the requirements of your country or institution) were obtained?
    \item[] Answer: \answerNo{} 
    \item[] Justification: No crowdsourcing experiments or research involving human subjects were conducted.
    \item[] Guidelines:
    \begin{itemize}
        \item The answer NA means that the paper does not involve crowdsourcing nor research with human subjects.
        \item Depending on the country in which research is conducted, IRB approval (or equivalent) may be required for any human subjects research. If you obtained IRB approval, you should clearly state this in the paper. 
        \item We recognize that the procedures for this may vary significantly between institutions and locations, and we expect authors to adhere to the NeurIPS Code of Ethics and the guidelines for their institution. 
        \item For initial submissions, do not include any information that would break anonymity (if applicable), such as the institution conducting the review.
    \end{itemize}

\item {\bf Declaration of LLM usage}
    \item[] Question: Does the paper describe the usage of LLMs if it is an important, original, or non-standard component of the core methods in this research? Note that if the LLM is used only for writing, editing, or formatting purposes and does not impact the core methodology, scientific rigorousness, or originality of the research, declaration is not required.
    \item[] Answer: \answerNA{} 
    \item[] Justification: The core method development in this research does not involve LLMs as any important, original, or non-standard components.
    \item[] Guidelines:
    \begin{itemize}
        \item The answer NA means that the core method development in this research does not involve LLMs as any important, original, or non-standard components.
        \item Please refer to our LLM policy (\url{https://neurips.cc/Conferences/2025/LLM}) for what should or should not be described.
    \end{itemize}

\end{enumerate}

\end{document}